\documentclass[sigconf]{acmart}
\AtBeginDocument{%
  }

\setcopyright{acmlicensed}
\copyrightyear{2025}
\acmYear{2025}
\setcopyright{cc}
\setcctype{by-nc-nd}
\acmConference[KDD '25]{Proceedings of the 31st ACM SIGKDD Conference on Knowledge Discovery and Data Mining V.2}{August 3--7, 2025}{Toronto, ON, Canada}
\acmBooktitle{Proceedings of the 31st ACM SIGKDD Conference on Knowledge Discovery and Data Mining V.2 (KDD '25), August 3--7, 2025, Toronto, ON, Canada}
\acmDOI{10.1145/3711896.3737442}
\acmISBN{979-8-4007-1454-2/2025/08}

\usepackage{hyperref}
\usepackage{url}
\usepackage{graphicx}
\usepackage{xspace}
\usepackage{makecell}
\usepackage{array}
\usepackage{algorithmicx,algorithm}
\usepackage{multirow}
\usepackage[normalem]{ulem}
\usepackage{subcaption}
\usepackage{tabularx}
\usepackage{color}
\usepackage[utf8]{inputenc}
\usepackage[T1]{fontenc}
\usepackage{CJKutf8}
\usepackage{enumitem}
\usepackage{diagbox}
\usepackage[flushleft]{threeparttable}
\usepackage{booktabs}
\usepackage{algorithmicx,algorithm}
\usepackage{wrapfig}
\usepackage[flushleft]{threeparttable}

\usepackage{amssymb}

\begin{document}

\title{TSFM-Bench: A Comprehensive and Unified Benchmark of Foundation Models for Time Series Forecasting}

\settopmatter{authorsperrow=3}
\author{Zhe Li}
\affiliation{
  \institution{East China Normal University}
  \country{Shanghai, China}
}

\email{zheli@stu.ecnu.edu.cn}

\author{Xiangfei Qiu}
\affiliation{
  \institution{East China Normal University}
  \country{Shanghai, China}
}
\email{xfqiu@stu.ecnu.edu.cn}

\author{Peng Chen}
\affiliation{
  \institution{East China Normal University}
  \country{Shanghai, China}
}
\email{pchen@stu.ecnu.edu.cn}

\author{Yihang Wang}
\affiliation{
  \institution{East China Normal University}
  \country{Shanghai, China}
}
\email{yhwang@stu.ecnu.edu.cn}

\author{Hanyin Cheng}
\affiliation{
  \institution{East China Normal University}
  \country{Shanghai, China}
}
\email{hycheng@stu.ecnu.edu.cn}

\author{Yang Shu}
\affiliation{
  \institution{East China Normal University}
  \country{Shanghai, China}
}
\email{yshu@dase.ecnu.edu.cn}

\author{Jilin Hu}
\affiliation{
  \institution{East China Normal University}
  \country{Shanghai, China}
}
\email{jlhu@dase.ecnu.edu.cn}

\author{Chenjuan Guo}
\affiliation{
  \institution{East China Normal University}
  \country{Shanghai, China}
}
\email{cjguo@dase.ecnu.edu.cn}

\author{Aoying Zhou}
\affiliation{
  \institution{East China Normal University}
  \country{Shanghai, China}
}
\email{ayzhou@dase.ecnu.edu.cn}

\author{Christian S. Jensen}
\affiliation{
  \institution{Aalborg University}
  \country{Aalborg, Denmark}
}
\email{csj@cs.aau.dk}

\author{Bin Yang}
\authornote{Corresponding author.}
\affiliation{
  \institution{East China Normal University}
 \country{Shanghai, China}
}
 \affiliation{
   \institution{Aalborg University}
  \country{Aalborg, Denmark}
 }
\email{byang@cs.aau.dk}

\renewcommand{\shortauthors}{Zhe Li et al.}

\begin{abstract}
Time Series Forecasting (TSF) is key functionality in numerous fields, such as financial investment, weather services, and energy management. Although increasingly capable TSF methods occur, many of them require domain-specific data collection and model training and do not generalize well when applied in other domains. Time Series Foundation Models (TSFMs) that are pre-trained on massive heterogeneous time series data aim to overcome these limitations. The prospects for generalizability have spurred the development of a new generation of TSFMs. This study proposes a benchmark, \textit{TSFM-Bench}, to facilitate comprehensive and unified evaluation of TSFMs. \textit{TSFM-Bench} covers a wide range of TSFMs, including those based on large language models and those pre-trained on time series data. \textit{TSFM-Bench} supports multiple forecasting scenarios, including zero-shot, few-shot, and full-shot, enabling assessment across the full range of adaptation strategies. \textit{TSFM-Bench} also provides a standardized experimental protocols for critical evaluation processes such as dataset splitting, loading, normalization, and few-shot sampling, facilitating consistency and fairness. We report on an extensive evaluation of TSFMs across a diverse range of datasets spanning multiple domains and exhibiting varied statistical characteristics. Specifically, we identify pros and cons and inherent limitations of existing TSFMs, and we propose potential directions for new model designs. 

\end{abstract}



\maketitle

\newcommand\kddavailabilityurl{https://github.com/decisionintelligence/TSFM-Bench}

\ifdefempty{\kddavailabilityurl}{}{
\begingroup\small\noindent\raggedright\textbf{KDD Availability Link:}\\
The source code of this paper has been made publicly available at \url{\kddavailabilityurl}.
\endgroup
}

\section{Introduction}
Time Series Forecasting~(TSF) is core functionality across application domains, such as finance investment, weather services, and energy management~\cite{deng2024disentangling,deng2021st,qiu2025easytime,xu2025kdd,li2024cikm,gao2025ssdts, tian2025iclr,xu2023spatial,xu2023tme}. Predicting future states based on historical observations enables data-based decision making. Consequently, TSF is an active field of research, as evidenced by the continued emergence of forecasting methods. 
However, most existing TSF methods require data-specific training prior to their use for forecasting~\cite{wu2024autocts++, patchtst,wu2023autocts+,qiu2025comprehensive,liu2025rethinking,wu2025k2vae}. These data-specific models (denoted as specific models in this study for simplicity) often exhibit limited generalizability to, and suboptimal performance on, novel or unseen data. Thus, there has been a recent surge in the development of time series foundation models (TSFMs) that aim to overcome these limitations~\cite{liang2024foundation,moiral, timer, gpt4ts, AimTS, shentu2025towards}.

As TSFMs with diverse architectures and training paradigms continue to emerge, our understanding of their strengths and limitations remains nascent. Existing studies on understanding these TSFMs focus primarily on qualitative analyses and categorization~\cite{liang2024foundation, jin2024position}. For instance, one study ~\cite{jin2024position} proposes a framework for LLM-based time series analysis, highlighting key opportunities and challenges for future research and advocating increased interdisciplinary collaboration. Similarly, another study~\cite{liang2024foundation} offers a methodology centered classification, outlining critical elements of TSFMs, such as model architectures, pre-training techniques, and adaptation strategies. However, these studies often do not include a quantitative evaluation of the TSFMs, which is crucial to assessing performance. Quantitative analyses enable researchers to make informed decisions about model selection and improvements.

\begin{figure*}[t]
  \centering
  \includegraphics[width=1\linewidth]{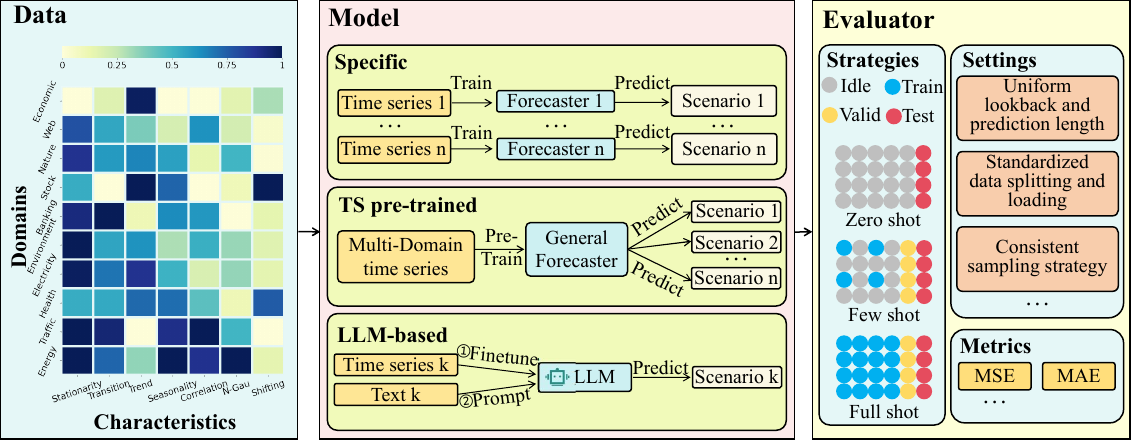}
  \caption{The \textit{TSFM-Bench} architecture with three core modules: Data, Model and Evaluator.} 
  \label{fig: architecture}
\end{figure*}

Further, different TSFMs often employ different training strategies, indicating that the performance of these models cannot be directly compared on a level playing field. As shown in Table~\ref{Comparison of experiment settings.}, taking few-shot scenario as an example, some TSFMs use uniform sampling~\cite{timer, units}, while others use front-end and back-end window sampling~\cite{timllm,S2IP-LLM}. And the sampling ratio and lookback lengths selected between models are also different.

\begin{table}[!t]
\caption{Comparison of experimental settings.}  
\small
\label{Comparison of experiment settings.}
  \resizebox{1.0\columnwidth}{!}{
    \begin{tabular}{c|l|c|c}    \toprule
    \multicolumn{1}{c|}{\textbf{Model}} & \textbf{Few-shot Sampling Type}  & \textbf{Sampling Ratio}  & \textbf{Lookback Length}\\
    \midrule
    Timer & {\color[HTML]{060607} Uniform window sampling} & 1--75\% & 672  \\
    UniTS & {\color[HTML]{060607} Uniform window sampling} & 5\%, 15\%, 20\% & 96  \\
    Time-LLM & {\color[HTML]{060607} Front-end window sampling} & 5\%, 10\% & 512  \\
    S$^{2}$IP-LLM & {\color[HTML]{060607} Front-end window sampling}& 5\%, 10\% & 512  \\
    \bottomrule
    \end{tabular}
  }
\end{table}

Comprehensive and unified benchmarks enable researchers to evaluate new models more rigorously, which is crucial for advancing the state-of-the-art~\cite{tfb,qiao2024class,qiu2025tab,tanmonash}. 
Most TSF benchmarks focus on evaluating the performance of data-specific models, while there are relatively few evaluations of TSFMs, as shown in Table ~\ref{Comparison between our work and other related benchmark.}. 
The only benchmark ProbTS~\cite{zhang2024probts} targeting TSFMs has two notable limitations: 1) the scope of TSFMs considered is not sufficiently comprehensive, as LLM-based models are ignored; 2) the supported evaluation strategies are too limited to fully reflect the performance of TSFMs, as few-shot scenarios are not considered.

Motivated by these observations, we present \textit{TSFM-Bench} that aims to facilitate fair and comprehensive empirical evaluation of TSFMs. 
First, \textit{TSFM-Bench} integrates multi-domain datasets with heterogeneous temporal characteristics to ensure the comprehensiveness of the assessment.
Second, \textit{TSFM-Bench} covers a variety of TSFMs, including LLM-based models and time series pre-trained models (TS pre-trained models). This is in addition to including state-of-the-art data-specific models to facilitate comparison with the TSFMs. 
Third, standardized experimental protocols facilitate fairness by enforcing unified dataset splitting, loading, normalization, and few-shot sampling strategies. 
Fourth, multiple evaluation strategies employ zero-shot, few-shot, and full-shot scenarios, paired with a variety of evaluation metrics. 
These properties combine to yield a fair and complete benchmark, enabling thorough evaluations with findings that are comparable.

In summary, we make the following main contributions:
\begin{itemize}[left=0.1cm]
\item  \textbf{Diversified models and datasets:} \textit{TSFM-Bench} covers state-of-the-art TSFMs, including LLM-based and TS pre-trained models. Additionally, it features comprehensive datasets that cover a wide range of domains and characteristics.

\item  \textbf{Comprehensive and fair evaluation strategies and settings:} \textit{TSFM-Bench} integrates zero-shot, few-shot, and full-shot scenarios, facilitating improved evaluation of model performance. In addition, it provides a unified experimental settings that standardizes dataset splitting, loading, and few-shot sampling, thereby facilitating fair comparisons of models.

\item  \textbf{In-depth quantitative analysis and insights:} Employing \textit{TSFM-Bench}, we report on extensive experiments covering different TSFMs. This way, we systematically evaluate the strengths and limitations of the TSFMs, providing valuable insights for the future design and optimization of such models.

\end{itemize}

\begin{table}[t]
  \centering
  \caption{Comparison between \textit{TSFM-Bench} and other time series forecasting benchmarks.}
  \resizebox{1.0\columnwidth}{!}{
    \begin{tabular}{c|ccc|ccc}
    \midrule
    \multirow{3}{*}{\textbf{\makecell{Time Series\\ Forecasting\\ Benchmarks}}} & \multicolumn{3}{c|}{\textbf{Evaluation Models}} & \multicolumn{3}{c}{\textbf{Evaluation Strategies}} \\
\cmidrule{2-7}    \multicolumn{1}{c|}{} & \multicolumn{1}{l}{\textbf{\makecell{LLM-based \\ models}}} & \multicolumn{1}{l}{\textbf{\makecell{TS pre-trained \\models}}} & \multicolumn{1}{l|}{\textbf{\makecell{Specific \\ models}}} & \multicolumn{1}{l}{\textbf{Zero-shot}} & \multicolumn{1}{l}{\textbf{Few-shot}} & \multicolumn{1}{l}{\textbf{Full-shot}}   \\
    \midrule
    \multicolumn{1}{c|}{M3~\cite{makridakis2000m3}} &$\times$ &$\times$ & $\surd$ &$\times$ &$\times$  & $\surd$  \\
    \multicolumn{1}{c|}{M4~\cite{makridakis2018m4}} &$\times$ &$\times$ & $\surd$ &$\times$ &$\times$  & $\surd$ \\
    \multicolumn{1}{c|}{LTSF-Liner~\cite{zeng2023transformers}}&$\times$ &$\times$ & $\surd$ &$\times$ &$\times$  & $\surd$   \\
    \multicolumn{1}{c|}{BasicTS~\cite{liang2022basicts}} &$\times$ &$\times$ & $\surd$ &$\times$ &$\times$  & $\surd$   \\
    \multicolumn{1}{c|}{BasicTS+~\cite{shao2023exploring}} &$\times$ &$\times$ & $\surd$ &$\times$ &$\times$  & $\surd$  \\
    \multicolumn{1}{c|}{Monash~\cite{godahewa2021monash}}&$\times$ &$\times$ & $\surd$ &$\times$ &$\times$  & $\surd$   \\
    \multicolumn{1}{c|}{Libra~\cite{bauer2021libra}} &$\times$ &$\times$ & $\surd$ &$\times$ &$\times$  & $\surd$ \\
    \multicolumn{1}{c|}{ProbTS~\cite{zhang2024probts}} &$\times$ &$\surd$ &$\surd$ & $\surd$ &$\times$  & $\surd$ \\
    \multicolumn{1}{c|}{TSLib~\cite{wang2024deep}} &$\times$ &$\times$ & $\surd$ &$\times$ &$\times$  & $\surd$ \\
    \multicolumn{1}{c|}{TFB~\cite{tfb}} &$\times$ &$\times$ & $\surd$ &$\times$ &$\times$  & $\surd$  \\
    \multicolumn{1}{c|}{\textit{TSFM-Bench} (ours)}& $\surd$ &$\surd$ & $\surd$ & $\surd$ &$\surd$ &$\surd$ \\
    \bottomrule
    \end{tabular}}
  \label{Comparison between our work and other related benchmark.} 
\end{table}

\section{Related Work}
\subsection{Time Series Forecasting}
In recent years, time series analysis has seen remarkable progress, with key tasks such as anomaly detection~\cite{D3R, liu2024elephant, miao2025parameter, hu2024multirc, wu2024catch}, classification~\cite{DBLP:conf/icde/YaoJC0GW24,DBLP:journals/pacmmod/0002Z0KGJ23}, and imputation~\cite{gao2025ssdts,escmtifs,wang2024spot,yu2025ginarp}, among others~\cite{wang2024entire,miao2024less,liu2025timecma,DBLP:conf/nips/HuangSZDWZW23,DBLP:journals/pvldb/YaoLJ00CW0G23}, gaining attention. Among these, Time Series Forecasting (TSF) is a crucial and widely studied task.
TSF models can be categorized into \textbf{data-specific models} (denoted ad \textbf{specific models} in this study for simplicity) and \textbf{Time series foundation models (TSFMs)}, depending on the scope and nature of their training data.

The former typically require training on domain-specific datasets and inference on corresponding test datasets. These models can be further classified into three main types:
\textbf{Statistical learning models}, while theoretically robust, often fail to capture nonlinear trends, resulting in limited predictive accuracy~\cite{ARIMA,hyndman2008forecasting}.  
\textbf{Machine learning models} are better at capturing non-linear relationships and complex temporal patterns, but often require manual feature engineering and careful model design~\cite{chen2016xgboost, friedman2001greedy}. 
\textbf{Deep learning models} leverage the representation learning capabilities of neural networks with rich datasets, and frequently achieve superior predictive performance compared to the other two approaches~\cite{dai2024periodicity, zhou2021informer,qiu2025duet,wang2023accurate,wanglabel,wang2023accurate}. 
Especially, patching operations, initially introduced by Triformer~\cite{cirstea2022triformer}, significantly enhance models' capabilities to capture temporal dependencies at the patch level. And PatchTST~\cite{patchtst} lies in establishing the fundamental role of patching operations in TSF tasks, with this breakthrough significantly enhancing the predictive accuracy of deep learning models.
However, all these models are limited by their strong coupling of training and inference data. They may not perform well when applied to unseen or out-of-distribution data.

TSFMs~\cite{liang2024foundation, lightgts, shentu2025towards, AimTS} can be divided into two sections: \textbf{TS pre-trained models} and \textbf{LLM-based models}. 
TS pre-trained models~\cite{moiral, timer, units, ansari2024chronos, rose}, pre-trained on diverse multi-domain time series datasets, exhibit generalization capabilities to perform excellent performance on TSF tasks with limited training data, such as few-shot scenario or even zero-shot scenario.
LLM-based models~\cite{gpt4ts, S2IP-LLM, timllm, unitime} refers to using the large language model (LLM), pre-trained with large-scale text data, in the TSF task. With the vast language understanding and context processing capabilities of LLM, these models demonstrate superior forecasting performance when handling previously unseen data. 

\subsection{Time Series Forecasting Benchmarks}
Several benchmarks have recently been proposed for TSF, as shown in Table~\ref{Comparison between our work and other related benchmark.}. However, their inherent limitations render it impossible to conduct a comprehensive and fair comparison between TSFMs and data-specific models for TSF tasks.

First, most benchmarks focus exclusively on data-specific models and overlook TSFMs~\cite{makridakis2000m3, makridakis2018m4}. The only exception is ProbTS~\cite{zhang2024probts}, which covers TS pre-trained models but excludes LLM-based models.
Given the impressive capabilities offered by TSFMs, such as zero-shot prediction significantly reducing the computational cost challenges associated with specific models, it is necessary to conduct a fair and comprehensive comparison between the two.

Second, current benchmarks lack support diverse evaluation strategies. Most TSF benchmarks disregard emerging features like zero-shot and few-shot scenarios, focusing instead on full-shot scenarios. 
Few-shot learning enables models to leverage small amounts of relevant data to fine-tune their performance. This capability not only enhances accuracy but also increases a model's flexibility at adapting to new tasks, making it more effective in dynamic environments. More importantly, the absence of standardized sampling methods for few-shot prediction compromises fair comparison. Thus, there is a pressing need for a comprehensive and fair evaluation strategies and settings to advance the development of TSFMs.

\textit{TSFM-Bench} is designed to be a reliable, comprehensive, and user-friendly benchmark, featuring a wider range of TSFMs and evaluation strategies. And it offers a unified experimental setting, ensuring consistent evaluations within a robust framework.

\section{TSFM-Bench}

To facilitate the evaluation and comparison of TSFMs, we propose \textit{TSFM-Bench}, a unified benchmark for TSFMs. Figure \ref{fig: architecture} shows its three core modules: Data, Model, and Evaluator. The data module includes time series datasets from different domains and with diverse characteristics, providing comprehensive data support for downstream TSF tasks. The model module includes TSF models, including specific models trained on domain-specific datasets, TS pre-trained models pre-trained with multi-domain time series datasets and LLM-based models pre-trained with large-scale text. The evaluation module offers standardized evaluation environment with comprehensive strategies and consistent settings, ensuring fair comparisons of models and facilitating reliable results.
\begin{table*}[t]
  \centering
  \caption{The statistics of evaluation datasets.}
  \label{tab: Evaluation Datasets}
  \resizebox{1\linewidth}{!}{
    \begin{tabular}{l|c|c|c|c|c|c|c|c|c|c|c|c}
    \toprule
    \textbf{Dataset} & \textbf{Variables} & \textbf{Timestamps} & \textbf{Split Ratio} & \textbf{Domain} & \textbf{Frequency} & \textbf{Seasonality} & \textbf{Trend} & \textbf{Stationarity} & \textbf{Transition} & \textbf{Shifting} & \textbf{Correlation} & \textbf{Non-Gaussianity} \\ \midrule
ETTh1 & 7 & 14,400 & 6:2:2 & Electricity & 1 hour & 0.731 & 0.853 & 0.001 & 0.020 & 0.061 & 0.630 & 0.104 \\
ETTh2 & 7 & 14,400 & 6:2:2 & Electricity & 1 hour & 0.615 & 0.906 & 0.022 & 0.042 & 0.404 & 0.509 & 0.150 \\
ETTm1 & 7 & 57,600 & 6:2:2 & Electricity & 15 mins & 0.760 & 0.614 & 0.000 & 0.027 & 0.063 & 0.612 & 0.134 \\
ETTm2 & 7 & 57,600 & 6:2:2 & Electricity & 15 mins & 0.615 & 0.906 & 0.003 & 0.038 & 0.406 & 0.504 & 0.174 \\
Electricity & 321 & 26,304 & 7:1:2 & Electricity & 1 hour & 0.945 & 0.795 & 0.005 & 0.010 & 0.075 & 0.802 & 0.120 \\
Traffic & 862 & 17,544 & 7:1:2 & Traffic & 1 hour & 0.880 & 0.263 & 0.000 & 0.011 & 0.067 & 0.814 & 0.186 \\
PEMS08 & 170 & 17,856 & 6:2:2 & Traffic & 5 mins & 0.850 & 0.112 & 0.000 & 0.005 & 0.025 & 0.807 & 0.083 \\
Solar & 137 & 52,560 & 6:2:2 & Energy & 10 mins & 0.919 & 0.478 & 0.000 & 0.025 & 0.198 & 0.785 & 0.342 \\
Wind & 7 & 48,673 & 7:1:2 & Energy & 15 mins & 0.557 & 0.639 & 0.007 & 0.028 & 0.132 & 0.507 & 0.200 \\
Weather & 21 & 52,696 & 7:1:2 & Environment & 10 mins & 0.652 & 0.649 & 0.000 & 0.037 & 0.214 & 0.694 & 0.134 \\
AQShunyi & 11 & 35,064 & 7:1:2 & Environment & 1 hour & 0.145 & 0.081 & 0.000 & 0.032 & 0.019 & 0.613 & 0.152 \\
Exchange & 8 & 7,588 & 7:1:2 & Economic & 1 day & 0.540 & 0.879 & 0.360 & 0.062 & 0.325 & 0.565 & 0.081 \\
FRED-MD & 107 & 728 & 7:1:2 & Economic & 1 month & 0.567 & 0.906 & 0.574 & 0.114 & 0.394 & 0.660 & 0.080 \\
ZafNoo & 11 & 19,225 & 7:1:2 & Nature & 30 mins & 0.757 & 0.670 & 0.043 & 0.034 & 0.078 & 0.599 & 0.186 \\
CzenLan & 11 & 19,934 & 7:1:2 & Nature & 30 mins & 0.764 & 0.685 & 0.160 & 0.053 & 0.158 & 0.683 & 0.178 \\
ILI & 7 & 966 & 7:1:2 & Health & 1 week & 0.809 & 0.714 & 0.169 & 0.038 & 0.721 & 0.674 & 0.064 \\
Covid-19 & 948 & 1,392 & 7:1:2 & Health & 1 day & 0.589 & 0.751 & 0.323 & 0.126 & 0.236 & 0.604 & 0.142 \\
NASDAQ & 5 & 1,244 & 7:1:2 & Stock & 1 day & 0.819 & 0.890 & 0.169 & 0.074 & 0.932 & 0.564 & 0.063 \\
NYSE & 5 & 1,243 & 7:1:2 & Stock & 1 day & 0.739 & 0.898 & 0.679 & 0.167 & 0.620 & 0.613 & 0.072 \\
NN5 & 111 & 791 & 7:1:2 & Banking & 1 day & 0.781 & 0.329 & 0.029 & 0.007 & 0.195 & 0.713 & 0.032 \\
Wike2000 & 2,000 & 792 & 7:1:2 & Web & 1 day & 0.617 & 0.501 & 0.078 & 0.037 & 0.104 & 0.719 & 0.097 \\ 
    \bottomrule
    \end{tabular}}
\end{table*}
\subsection{Data}

High-quality and diverse time series datasets enable a comprehensive evaluation of model performance across various conditions, thereby facilitating the identification and selection of models that are best suited for specific forecasting scenarios and real-world applications.
The dataset provides broad coverage across domains and diverse statistical characteristics, enabling a more comprehensive comparison of models in terms of both predictive accuracy and generalization ability.
Table~\ref{tab: Evaluation Datasets} lists statistics of the 21 multivariate time series datasets. 

(1) \textbf{Domains}:To ensure a thorough and representative evaluation of model performance across diverse real-world scenarios, we compile a collection of datasets spanning ten distinct domains. Specifically, we include ETT~\cite{zhou2021informer} and Electricity~\cite{misc_electricityloaddiagrams20112014_321} from the \textit{electricity} domian, 
Traffic~\cite{wu2021autoformer} and PEMS08~\cite{song2020spatial} from the \textit{traffic} domain, 
Solar~\cite{lai2018modeling} and Wind~\cite{li2022generative} from the \textit{energy} domain, 
Weather~\cite{wu2021autoformer} and AQShunyi~\cite{zhang2017cautionary} from the \textit{environment} domain, 
Exchange~\cite{lai2018modeling} and FRED-MD~\cite{mccracken2016fred} from the \textit{economics} domain, 
ZafNoo and CzeLan~\cite{poyatos2020global} from the \textit{nature} domain, 
ILI~\cite{wu2021autoformer} and Covid-19~\cite{panagopoulos2021transfer} from the \textit{health} domain, 
NASDAQ and NYSE~\cite{feng2019temporal} from the \textit{stock  market} domain, 
NN5~\cite{taieb2012review} from the \textit{banking} domain 
and Wike2000~\cite{gasthaus2019probabilistic} from the \textit{web} domain. 
Additional information about the datasets can be found in the Appendix~\ref{Datasets Collection}. 

(2) \textbf{Characteristics}: 
Our comprehensive analysis encompasses a diverse set of time series characteristics that capture various aspects of temporal dynamics, including seasonality, trend, stationarity, transition, shifting, correlation, and non-Gaussianity~\cite{tfb,zhang2024probts}.
\textit{Seasonality} refers to repeating patterns or cycles at regular intervals, such as daily, weekly, or yearly variations.
\textit{Trend} captures the long-term underlying direction of a time series, reflecting persistent upward, downward, or stable movements over time. 
\textit{Stationarity} reflects the statistical properties of a time series, such as mean and variance, which do not change over time. 
\textit{Transition} represents sudden or gradual shifts in a time series. 
\textit{Shifting} refers to changes in the level or timing of the data and includes vertical and horizontal shifts. 
\textit{Correlation} represents the relationship or dependence among different channels. 
\textit{Non-Gaussianity (N-Gau)} represents deviations from normal distribution, often exhibiting skewness or kurtosis. 
The ``Data'' part of Figure \ref{fig: architecture} shows data domains with varying characteristic distributions. This facilitates comprehensive evaluation of prediction accuracy and generalization cap abilities under varying data characteristics. 
The formulas used to calculate these characteristics are provided in Appendix~\ref{characteristic formula}.

\subsection{Model}

\subsubsection{Time Series Data-Specific Models (Specific models)}
\textbf{Specific models} typically require training on specific datasets and perform inference on the corresponding datasets. To better showcase the capabilities of TSFMs, we select several SOTA data-specific models for comparison. We include: 1) CNN-based models: TimesNet~(TsNet)~\cite{TimesNet}, which treat time series as sequences of vectors and leverage CNNs to capture temporal dependencies. 2) Transformer-based models: FEDformer~(FedF)~\cite{zhou2022fedformer}, iTransformer~(iTrans)~\cite{liu2023itransformer}, and PatchTST~(Patch)~\cite{patchtst}, which are capable of capturing more complex temporal dynamics, leading to significantly improved forecasting performance. 3) MLP-based models: FITS~\cite{xu2024fitsmodelingtimeseries}, TimeMixer\\~(TMixer)~\cite{wang2024timemixer}, and DLinear~(Dlin)~\cite{zeng2023transformers}, with their simple architecture and relatively few parameters, have demonstrated strong forecasting accuracy as well.

\subsubsection{Time Series Foundation Models (TSFMs)} 
According to the pre-training data type, TSFMs can be classified into two categories:

\textbf{TS pre-trained models}: Pre-training on multi-domain time series datasets has gained significant attention in recent years.  We incorporate TS pre-trained models into \textit{TSFM-Bench}, categorizing them into four types based on the pre-training approach: reconstruction, autoregressive, direct prediction, and hybrid training. 
1) Reconstruction methods: MOIRAI~\cite{moiral}, UniTS~\cite{units}, Moment~\cite{moment} restore the features of time series data, enabling them to extract valuable information in an unsupervised manner. This type of method mainly adopts the encoder architecture.
2) Autoregressive methods: TimesFM~\cite{timesfm}, Timer~\cite{timer}, employ next token prediction to learn time series representation. This type of method mainly adopts the decoder architecture. 
3) Direct prediction methods: TTM~\cite{ekambaram2024ttms}, unify the training process between pre-training and downstream tasks, allowing models to exhibit strong adaptability when transitioning to downstream forecasting tasks. 
4) Hybrid pre-training methods: ROSE~\cite{rose} and Chronos~\cite{ansari2024chronos}, combines the strengths of both reconstruction and direct prediction to learn generalized time series representations.

\textbf{LLM-based models}: LLMs-based methods leverage the strong representational capacity and sequential modeling capability of LLM to capture complex patterns in time series data. To more comprehensively evaluate the TSFMs, we incorporate existing LLM-based models into \textit{TSFM-Bench}, focusing primarily on parameter-efficient fine-tuning and prompting: 
1) Parameter-efficient fine-tuning methods: GPT4TS~\cite{gpt4ts} and CALF~\cite{liu2024taming} selectively adjust specific parameters such as positional encoding and layer normalization of LLM, enabling the model to quickly adapt to time series while retaining most of pre-trained knowledge. 
2) Prompting methods: LLMMixer~\cite{kowsher2024llm}, UniTime~\cite{unitime}, S$^{2}$IP-LLM~\cite{S2IP-LLM} and Time-LLM~\cite{timllm}, focus on designing prompts, such as learnable prompts, prompt pools, and domain-specific instructions to activate time series knowledge in LLM.

\subsection{Evaluator}
To ensure a fair and comprehensive evaluation of the performance of various TSFMs, we standardize the evaluation in three key areas: strategies, settings and metrics.

\subsubsection{Strategies}
Considering that current benchmarks typically adopt a single evaluation approach, focusing only on a zero-shot or full-shot scenario, this limits the ability to comprehensively assess prediction performance. We propose a more comprehensive quantitative evaluation that offers researchers a broader understanding under different conditions, including zero-shot, few-shot, and full-shot scenarios. As shown in Figure \ref{fig: architecture}, we divide the downstream evaluation data into training, validation and test datasets.

(1) The zero-shot scenario only uses the test dataset to evaluate the generalization ability of TSFMs to new datasets, assessing whether the model has truly learned general knowledge from vast amounts of pre-training data. 

(2) The few-shot scenario utilizes a subset of training dataset in a ratio and full validation dataset for fine-tuning, reflecting the prediction performance in low-data learning scenarios. This approach assesses whether models can effectively generalize and reason with minimal data support. 

(3) The full-shot scenario utilizes full training dataset and validation dataset for fine-tuning. Evaluates performance when using all available data, revealing its upper bound performance.

\begin{table}[t]
\caption{Comparison of different sampling strategies.}  
\small
\label{Comparison of different sampling methods}
  \resizebox{1\columnwidth}{!}{
\begin{tabular}{c|c|llllll} \midrule
\multicolumn{2}{c|}{\multirow{1}{*}{\textbf{ETTm1}}} & \multicolumn{2}{c}{\textbf{TTM}} & \multicolumn{2}{c}{\textbf{Timer}} & \multicolumn{2}{c}{\textbf{UniTS}} \\
\multicolumn{1}{c}{\textbf{Strategy}} & \multicolumn{1}{c|}{\textbf{Format}} & \multicolumn{1}{c}{\textbf{MAE}} & \multicolumn{1}{c}{\textbf{MSE}} & \multicolumn{1}{c}{\textbf{MAE}} & \multicolumn{1}{c}{\textbf{MSE}} & \multicolumn{1}{c}{\textbf{MAE}} & \multicolumn{1}{c}{\textbf{MSE}} \\\midrule
\textbf{Random} & \textbf{Window sample$^1$} & 0.352 & 0.311 & \textbf{0.343} & \textbf{0.288} & 0.426 & 0.419 \\
\addlinespace\cline{1-8} \addlinespace
\textbf{Uniform} & \textbf{Window sample} & 0.347 & 0.303 & \textbf{0.342} & \textbf{0.286} & 0.387 & 0.358 \\
\addlinespace\cline{1-8} \addlinespace
\multirow{2}{*}{\textbf{Front-end}} & \textbf{Window sample} & \textbf{0.374} & \textbf{0.344} & 0.476 & 0.536 & 0.477 & 0.639 \\\addlinespace\cline{2-8} \addlinespace
 & \textbf{Point sample$^2$} & \textbf{0.393} & \textbf{0.366} & 0.510 & 0.598 & 0.510 & 0.598 \\
 \addlinespace\cline{1-8} \addlinespace
\multirow{2}{*}{\textbf{Back-end}}  & \textbf{Window sample}& \textbf{0.356} & \textbf{0.310} & 0.382 & 0.334 & 0.385 & 0.368 \\\addlinespace\cline{2-8} \addlinespace
  & \textbf{Window sample} & \textbf{0.363} & \textbf{0.318} & 0.386 & 0.340 & 0.386 & 0.371
  \\\bottomrule  \end{tabular}}
\begin{tablenotes}
\fontsize{6}{2}\selectfont
\item[1] Window sample refers to first dividing the dataset into windows (lookback length + prediction length), and then selecting a specified proportion of these window samples.
  \item[2] Point sample refers to directly extracting a specified proportion of data points from the dataset.
  \item[3] To ensure the temporal continuity of a sample, point sampling is only applicable to the front-end and back-end sampling strategies.
\end{tablenotes}
\end{table}

\subsubsection{Settings}
Different evaluation settings can cause significant discrepancies in model performance, leading to unfair comparisons of their actual capabilities. To address this, we standardize the settings, including lookback and prediction lengths, data spliting and loading, and consistent sampling strategies. 

(1) \textbf{Uniform lookback and prediction lengths}: The lookback length determines the amount of historical information the model receives, and different lengths lead to varying prediction results. Following the common practices, for short-term forecasting (e.g., ILI, NASDAQ, NN5 and Wike2000), we use four prediction lengths of 24, 36, 48, 60 and two lookback lengths of 36, 104. For long-term forecasting (e.g., ETTh1, Traffic and others), we use four prediction lengths of 96, 192, 336, 720 and three lookback lengths of 96, 336, 512. 
For each prediction length, we report the best performance across different lookback lengths.

(2) \textbf{Standardized data splitting and loading}: We standardize the division of the training, validation and test datasets, as well as the partitioning of each time series sample for all models. To ensure that different models use a consistent test length, we do not apply the ``Drop Last'' operation during testing~\cite{tfb,qiu2025tab}.

(3) \textbf{Consistent sampling strategies}: 
We integrate various sampling strategies, including random sampling, uniform sampling, front-end sampling, and back-end sampling. In addition, we support both window and point sampling formats --- see Table~\ref{Comparison of different sampling methods}. The results demonstrate that different sampling types significantly impact model performance, even leading to substantial performance gaps. This indicates that data sampling types play a crucial role in few-shot learning, and standardized experimental settings are essential to fairly evaluate the actual performance of TSFMs. By default, we consistently use 5\% uniform window sampling across all models for assessment and reporting to ensure a fair comparison. Our framework supports the seamless transition to other strategies.

\subsubsection{Metrics}
We incorporate a variety of metrics for evaluation, including Mean Absolute Error (MAE) and Mean Squared Error (MSE), among others. Different metrics offer a multifaceted evaluation of model performance, each providing unique insights from different perspectives.

\section{Experiments}
\subsection{Benchmarking results}

\subsubsection{Zero-shot Evaluation}
Specific models typically require training on domain-relevant datasets, while most LLM-based models necessitate fine-tuning of the LLM backbones or some additional components to achieve optimal performance on downstream tasks.
Thus, in zero-shot evaluation, we focus on TS pre-trained models that can perform forecasting in a zero-shot scenario. The zero-shot performance of these TS pre-trained models are present in Table~\ref{tab:zero}. The main findings are as follows: 

(1) It is evident that no single model consistently outperforms others across all datasets. The observation may be attributed to differences in architecture, scale, pre-training datasets, and objectives across current TS pre-trained models.

(2) From the Table~\ref{tab:few} and Table~\ref{full-shot}, TS pre-trained models achieve superior zero-shot performance compared to the few-shot and full-shot results of specific models across most datasets. This indicates that TSFMs pre-trained on multi-domain time series data possess strong generalization capabilities. Their impressive performance in zero-shot scenario suggests promising potential for TSF in new domains and tasks.

(3) It is noteworthy that TS pre-trained models exhibit suboptimal zero-shot performance in energy, economic, and electricity domains compared to specific models under full-shot scenario. 
These observations suggest that TSFMs require more diverse and comprehensive pre-training datasets to enhance their generalization across heterogeneous scenarios.
\begin{table}[t]
\centering
\small
\caption{TS Pre-trained model results in the \textbf{zero-shot scenario}. The results are the average MSE of all prediction lengths. The complete results of MAE and MSE can be found in Appendix~\ref{Full results}.}
\resizebox{1\columnwidth}{!}{
\begin{tabular}{c|ccccccc}
\toprule
\textbf{Dataset} & \textbf{Chronos} & \textbf{TimesFM} & \textbf{UniTS} & \textbf{MOIRAI} & \textbf{TTM} & \textbf{ROSE} & \textbf{Timer} \\
\midrule
ETTh1 & 0.445 & 0.451 & 0.527 & 0.433 & \textbf{0.394} & \uline{0.401} & 0.432 \\
ETTh2 & 0.368 & 0.403 & 0.405 & 0.360 & \textbf{0.345} & \uline{0.346} & 0.360 \\
ETTm1 & \textbf{0.425} & \uline{0.429} & 0.713 & 0.517 & 0.692 & 0.525 & 0.666 \\
ETTm2 & \textbf{0.292} & 0.335 & 0.321 & 0.340 & 0.333 & 0.299 & \uline{0.293} \\
Electricity & \uline{0.166} & \textbf{0.155} & 0.432 & 0.202 & 0.205 & 0.234 & 0.181 \\
Traffic & 0.448 & \textbf{0.370} & 1.015 & \uline{0.380} & 0.564 & 0.588 & 0.451 \\
PEMS08 & \uline{0.697} & 1.485 & 1.252 & \textbf{0.241} & 1.730 & 1.369 & 0.866 \\
Solar & \textbf{0.422} & \uline{0.438} & 0.875 & 0.944 & 0.815 & 0.505 & 0.651 \\
Wind & 1.371 & 1.613 & 1.425 & \uline{1.230} & 1.337 & 1.251 & \textbf{1.201} \\
Weather & 0.266 & \textbf{0.231} & 0.291 & 0.308 & 0.265 & 0.274 & \uline{0.265} \\
AQShunyi & 0.788 & 0.871 & 0.876 & \textbf{0.680} & 0.715 & 0.734 & \uline{0.707} \\
Exchange & 0.489 & 0.433 & \uline{0.424} & 0.452 & \textbf{0.390} & 0.618 & 0.501 \\
FRED-MD & \uline{64.290} & \textbf{58.121} & 127.798 & 77.887 & - & - & - \\
ZafNoo & 0.556 & 0.638 & 0.715 & \textbf{0.527} & 0.690 & \uline{0.545} & 0.549 \\
CzeLan & \textbf{0.315} & \uline{0.317} & 0.735 & 0.647 & 0.698 & 0.455 & 0.492 \\
ILI & \uline{2.865} & \textbf{1.811} & 4.066 & 3.414 & - & - & 3.721 \\
Covid-19 & \textbf{1.457} & 13.275 & \uline{3.416} & 5.673 & - & - & 6.836 \\
NASDAQ & \textbf{0.883} & 1.034 & 1.105 & 1.081 & - & - & \uline{0.934} \\
NYSE & \textbf{0.589} & \uline{0.623} & 1.219 & 1.553 & - & - & 0.988 \\
NN5 & 1.113 & \textbf{0.780} & 1.261 & \uline{0.787} & - & - & - \\
Wike2000 & \uline{511.873} & \textbf{477.539} & 684.147 & - & - & - & 823.995 \\
\bottomrule
\end{tabular}

}
\begin{tablenotes}
\fontsize{6}{2}\selectfont
\item Due to limitations in the time series splitting strategy or model training strategy used by some models, they are unable to adapt to short-term prediction datasets, which is shown with --.
\end{tablenotes}
\label{tab:zero}
\end{table}

\begin{table*}[t]
\centering
\caption{Model results in the \textbf{5\% few-shot scenario}. The results are the average MSE of all prediction lengths. The complete few-shot results of MAE and MSE can be found in Appendix~\ref{Full results}.}  
\resizebox{1\linewidth}{!}{
\begin{tabular}{c|cccccccc|cccccc|cccccccc}
\toprule
\multicolumn{1}{c|}{\textbf{Dataset}} & \textbf{Chronos} & \textbf{TimesFM} & \textbf{Moment} & \textbf{UniTS} & \textbf{MOIRAI} & \textbf{TTM} & \textbf{ROSE} & \textbf{Timer} & \textbf{LLMMixer} & \textbf{S$^2$IPLLM} & \textbf{CALF} & \textbf{Time-LLM} & \textbf{GPT4TS} & \textbf{UniTime} & \textbf{Patch} & \textbf{Dlin} & \textbf{FITS} & \textbf{iTrans} & \textbf{FedF} & \textbf{TsNet} & \textbf{TMixer}  \\
\midrule
ETTh1 & 0.445 & 0.445 & 0.459 & 0.446 & 0.433 & \textbf{0.396} & \uline{0.405} & 0.408 & 0.524 & 0.674 & 0.444 & 0.674 & 0.467 & 0.799 & 0.447 & 0.476 & 0.781 & 0.572 & 0.557 & 0.855 & 0.638 \\
ETTh2 & 0.367 & 0.390 & 0.353 & 0.387 & 0.360 & \textbf{0.340} & \uline{0.348} & 0.375 & 0.381 & 0.396 & 0.365 & 0.396 & 0.376 & 0.418 & 0.386 & 0.500 & 0.443 & 0.415 & 0.477 & 0.449 & 0.442 \\
ETTm1 & 0.425 & 0.825 & \uline{0.358} & 0.424 & 0.517 & 0.359 & 0.381 & \textbf{0.356} & 0.369 & 0.373 & 0.372 & 0.373 & 0.389 & 0.409 & 0.374 & 0.381 & 0.505 & 0.439 & 0.666 & 0.483 & 0.554 \\
ETTm2 & 0.292 & 0.323 & \uline{0.260} & 0.300 & 0.340 & 0.261 & 0.265 & \textbf{0.256} & 0.266 & 0.270 & 0.271 & 0.270 & 0.287 & 0.274 & 0.277 & 0.296 & 0.265 & 0.307 & 0.418 & 0.323 & 0.312 \\
Electricity & \uline{0.166} & 0.235 & 0.191 & 0.211 & 0.232 & 0.184 & 0.222 & \textbf{0.161} & 0.185 & - & 0.185 & - & 0.208 & 0.202 & 0.216 & 0.179 & 0.492 & 0.216 & 0.302 & 0.262 & 0.580 \\
Traffic & 0.431 & 0.698 & 0.452 & \textbf{0.263} & \uline{0.380} & 0.497 & 0.573 & 0.397 & - & - & 0.610 & - & 0.433 & 0.433 & 0.433 & 0.426 & 1.216 & 0.513 & 0.836 & 0.828 & 0.475 \\
PEMS08 & 0.697 & 1.190 & 0.416 & 0.462 & \textbf{0.241} & 0.682 & 0.866 & \uline{0.344} & 0.392 & 0.450 & 0.481 & 0.416 & 0.466 & 0.419 & 0.457 & 0.440 & 1.272 & 0.503 & 2.419 & 0.462 & 0.575 \\
Solar & 0.422 & 0.825 & \uline{0.217} & 0.480 & 1.009 & 0.238 & 0.262 & \textbf{0.194} & 0.236 & - & 0.247 & - & 0.267 & 0.218 & 0.261 & 0.226 & 0.449 & 0.251 & 0.710 & 0.251 & 0.263 \\
Wind & 1.448 & 21.201 & 1.337 & 1.306 & 1.230 & 1.191 & 1.144 & \uline{1.135} & 1.338 & 1.338 & 1.328 & 1.321 & 1.347 & 1.358 & 1.137 & \textbf{1.100} & 1.151 & 1.215 & 1.366 & 1.238 & 1.282 \\
Weather & 0.266 & 0.280 & 0.240 & \uline{0.226} & 0.308 & \uline{0.226} & 0.237 & \textbf{0.226} & 0.236 & 0.237 & 0.238 & 0.237 & 0.252 & 0.294 & 0.245 & 0.254 & 0.254 & 0.269 & 0.354 & 0.277 & 0.237 \\
AQShunyi & 0.817 & 0.826 & 0.833 & 0.825 & \textbf{0.680} & \uline{0.706} & 0.718 & 0.722 & 0.852 & 0.901 & 0.832 & 0.859 & 0.722 & 0.904 & 0.723 & 0.732 & 0.841 & 0.807 & 0.817 & 0.787 & 0.911 \\
Exchange & \uline{0.374} & \textbf{0.296} & 0.484 & 0.406 & 0.452 & 0.398 & 0.621 & 0.520 & 0.444 & 0.407 & 0.421 & 0.407 & 0.531 & 0.458 & 0.429 & 0.381 & 0.399 & 0.416 & 0.643 & 0.427 & 0.433 \\
ZafNoo & 0.590 & 0.589 & 0.534 & 0.604 & 0.527 & \textbf{0.503} & 0.539 & 0.512 & 0.716 & 0.600 & 0.742 & 0.593 & 0.564 & 0.803 & 0.541 & \uline{0.508} & 0.550 & 0.617 & 0.683 & 0.595 & 0.538 \\
CzeLan & 0.644 & 0.712 & 0.366 & 0.307 & 0.647 & 0.811 & 0.303 & \textbf{0.212} & 0.362 & 0.613 & 0.287 & 0.319 & 0.431 & 0.401 & \uline{0.257} & 0.354 & 0.539 & 0.366 & 0.737 & 0.258 & 0.429 \\

\bottomrule
\end{tabular}
}
\begin{tablenotes}
\fontsize{5}{0}\selectfont
\item[1] With the need to meet the sampling requirements of 5\% training data under few-shot scenario, we only test on long-term prediction datasets.
\item[2] The maximum training duration is constrained to a maximum of 10 hours. Models that exceeded this threshold are represented with --.
\end{tablenotes}
\label{tab:few}
\end{table*}

\subsubsection{Few-shot Evaluation}
We assess TS pre-trained models, LLM-based models, and specific models with a 5\% few-shot scenario, and the results are reported in Table~\ref{tab:few}. The main findings are as follows: 

(1) TS pre-trained models generally outperform both LLM-based and specific models, with 13 out of 14 datasets showing a lead. 
This advantage likely stems from their ability to capture fundamental temporal patterns during pre-training, enabling rapid adaptation to downstream tasks and strong performance under data scarce conditions.

(2) The majority of LLM-based models perform worse than SOTA specific models. The text-based pre-training knowledge of LLM has shown limited utility in TSF tasks and may even interfere with the model's reasoning. Therefore, a key direction for future research lies in effectively leveraging LLM' pre-training knowledge by aligning the semantic spaces of text and time series data.

(3) A few specific models, such as DLinear, achieve strong performance in some datasets, potentially because their smaller parameter sizes allow for faster fitting of simple time series information. This indicates that research on few-shot learning may not only focus on TSFMs but also on some efficient small models.

\subsubsection{Full-shot Evaluation}
Since full-shot training in some TSFMs may take substantially long time, which
violates the original intention of TSFMs, we only select several representative TSFMs that are more efficient in training in the full-shot scenario. As shown in Table \ref{full-shot}:

(1) In terms of predictive accuracy, TS pre-trained models and specific models collectively outperform LLM-based models, with ROSE achieving best and suboptimal results on nearly one-third of the benchmark datasets. 
Notably, while all TS pre-trained models attain optimal performance on over half of the datasets, this comes at the cost of prohibitively high computational overhead during pre-training. This underscores the critical need for researchers to balance accuracy and efficiency in TSFMs development.

(2) Compared to the few-shot results in Table~\ref{tab:few}, a few TS pre-trained models show a decline in performance on some datasets when fine-tuned on more training datasets. 
That indicates that unlocking the full potential of TSFMs through scaling laws is still an essential challenge.

(3) Intriguingly, many large pre-trained models perform worse in few-shot and full-shot settings than in zero-shot. This paradox may arise from the limited diversity and scale of single-domain datasets, which constrain effective fine-tuning and lead to marginal gains despite high computational costs.

(4) LLM-based models perform better in the full-shot scenario compared to their performance in the few-shot scenario, likely because the increase in training data helps unlock time series related knowledge embedded in the LLM.

\begin{table*}[]
\centering
\caption{Model results in the \textbf{full-shot scenario}. The results are the average MSE of all prediction lengths. The complete full-shot results of MAE and MSE can be found in Appendix~\ref{Full results}.} 
\resizebox{1\linewidth}{!}{
\begin{tabular}{c|cccccc|cccc|ccccccc}
\toprule
\multicolumn{1}{c|}{\textbf{Dataset}} & \textbf{Chronos} & \textbf{UniTS} & \textbf{MOIRAI} & \textbf{TTM} & \textbf{ROSE} & \textbf{Timer} & \textbf{LLMMixer} & \textbf{CALF} & \textbf{GPT4TS} & \textbf{UniTime} & \textbf{Patch} & \textbf{Dlin} & \textbf{FITS} & \textbf{iTrans} & \textbf{FedF} & \textbf{TsNet} & \textbf{TMixer} \\
\midrule
ETTh1 & 0.445 & 0.475 & 0.433 & \uline{0.397} & \textbf{0.393} & 0.485 & 0.432 & 0.452 & 0.425 & 0.500 & 0.418 & 0.424 & 0.407 & 0.439 & 0.433 & 0.468 & 0.427 \\
ETTh2 & 0.367 & 0.408 & 0.360 & 0.346 & \textbf{0.334} & 0.401 & 0.367 & 0.394 & 0.355 & 0.370 & 0.351 & 0.470 & \uline{0.335} & 0.370 & 0.406 & 0.390 & 0.349 \\
ETTm1 & 0.425 & 0.384 & 0.517 & \textbf{0.345} & 0.351 & 0.431 & 0.363 & 0.388 & \uline{0.349} & 0.358 & 0.349 & 0.355 & 0.357 & 0.361 & 0.567 & 0.407 & 0.355 \\
ETTm2 & 0.292 & 0.313 & 0.340 & 0.258 & \textbf{0.250} & 0.289 & 0.265 & 0.282 & 0.268 & 0.268 & 0.256 & 0.259 & \uline{0.254} & 0.269 & 0.335 & 0.292 & 0.257 \\
Electricity & 0.166 & 0.166 & 0.259 & 0.178 & 0.164 & 0.168 & 0.367 & \textbf{0.163} & 0.168 & 0.173 & 0.171 & 0.167 & 0.169 & \uline{0.163} & 0.218 & 0.190 & 0.184 \\
Traffic & 0.431 & \textbf{0.390} & 0.399 & 0.668 & 0.414 & \uline{0.393} & - & 0.399 & 0.423 & 0.411 & 0.397 & 0.418 & 0.436 & 0.397 & 0.620 & 0.617 & 0.409 \\
PEMS08 & 1.081 & 0.608 & \textbf{0.241} & 0.687 & 0.364 & \uline{0.292} & 0.698 & 0.452 & 0.342 & 0.345 & 0.332 & 0.399 & 0.520 & 0.298 & 1.337 & 0.345 & 0.292 \\
Solar & 0.422 & \textbf{0.181} & 1.026 & 0.189 & \uline{0.185} & 0.225 & 0.258 & 0.210 & 0.304 & 0.192 & 0.200 & 0.224 & 0.232 & 0.202 & 0.641 & 0.211 & 0.193 \\
Wind & 1.448 & 1.243 & 1.230 & \uline{1.101} & 1.134 & 1.391 & 1.264 & 1.310 & 1.288 & 1.309 & 1.111 & \textbf{1.077} & 1.137 & 1.133 & 1.247 & 1.253 & 1.134 \\
Weather & 0.266 & 0.245 & 0.308 & 0.226 & \uline{0.225} & 0.271 & 0.227 & 0.240 & 0.233 & 0.230 & \textbf{0.224} & 0.242 & 0.243 & 0.232 & 0.312 & 0.255 & 0.226 \\
AQShunyi & 0.817 & 0.802 & \textbf{0.680} & \uline{0.698} & 0.726 & 0.787 & 0.803 & 0.831 & 0.819 & 0.830 & 0.703 & 0.706 & 0.713 & 0.706 & 0.763 & 0.727 & 0.706 \\
Exchange & 0.379 & 0.501 & 0.452 & 0.415 & 0.398 & 0.508 & 0.367 & 0.378 & 0.474 & 0.437 & 0.351 & \textbf{0.292} & \uline{0.349} & 0.360 & 0.501 & 0.406 & 0.381 \\
FRED-MD & \textbf{64.290} & 97.620 & 77.616 & - & - & - & 222.228 & \uline{73.515} & 99.082 & 83.918 & 78.174 & 96.092 & 124.065 & 75.609 & 115.966 & 81.046 & 76.156 \\
ZafNoo & 0.570 & 0.523 & 0.527 & \uline{0.503} & 0.512 & 0.646 & 0.527 & 0.543 & 0.516 & 0.520 & 0.508 & \textbf{0.495} & 0.522 & 0.522 & 0.575 & 0.536 & 0.517 \\
CzeLan & 0.644 & 0.249 & 0.647 & \textbf{0.206} & \uline{0.214} & 0.247 & 0.260 & 0.265 & 0.284 & 0.255 & 0.222 & 0.284 & 0.235 & 0.218 & 0.309 & 0.236 & 0.218 \\
ILI & 2.815 & 2.250 & 3.418 & - & - & 3.558 & 3.012 & 2.185 & 2.876 & 3.299 & 1.901 & 2.185 & 2.550 & \uline{1.857} & 2.485 & 2.163 & \textbf{1.819} \\
Covid-19 & \uline{1.457} & 3.193 & \textbf{0.241} & - & - & 2.942 & - & 1.524 & 1.837 & 1.781 & 1.607 & 8.074 & 5.669 & 1.488 & 2.678 & 2.609 & 12.941 \\
NASDAQ & \uline{0.936} & 1.180 & 1.081 & - & - & 0.972 & 1.242 & 1.026 & 1.179 & 1.099 & 0.972 & 1.400 & 1.044 & 0.944 & \textbf{0.933} & 0.954 & 1.005 \\
NYSE & 0.589 & 0.718 & 1.550 & - & - & 0.806 & 1.793 & 0.480 & 0.666 & 0.636 & 0.482 & 0.565 & 0.644 & \uline{0.458} & \textbf{0.405} & 0.498 & 0.666 \\
NN5 & 1.103 & 0.859 & 0.787 & - & - & - & 1.843 & 0.690 & 1.220 & 0.955 & 0.689 & 0.692 & 0.811 & \uline{0.660} & 0.691 & 0.684 & \textbf{0.655} \\
Wike2000 & \uline{511.873} & \textbf{492.262} & - & - & - & 533.506 & - & - & 535.125 & 545.181 & 513.964 & 630.769 & 732.541 & 545.647 & 733.552 & 527.789 & 518.679 \\
\bottomrule
\end{tabular}
}
\begin{tablenotes}
\fontsize{5}{0}\selectfont
\item The maximum training duration was constrained to a maximum of 10 hours. Models that exceeded this threshold are represented with --.  
\end{tablenotes}
\label{full-shot}
\end{table*}

\begin{figure}[t]
  \centering
\includegraphics[width=1\linewidth]{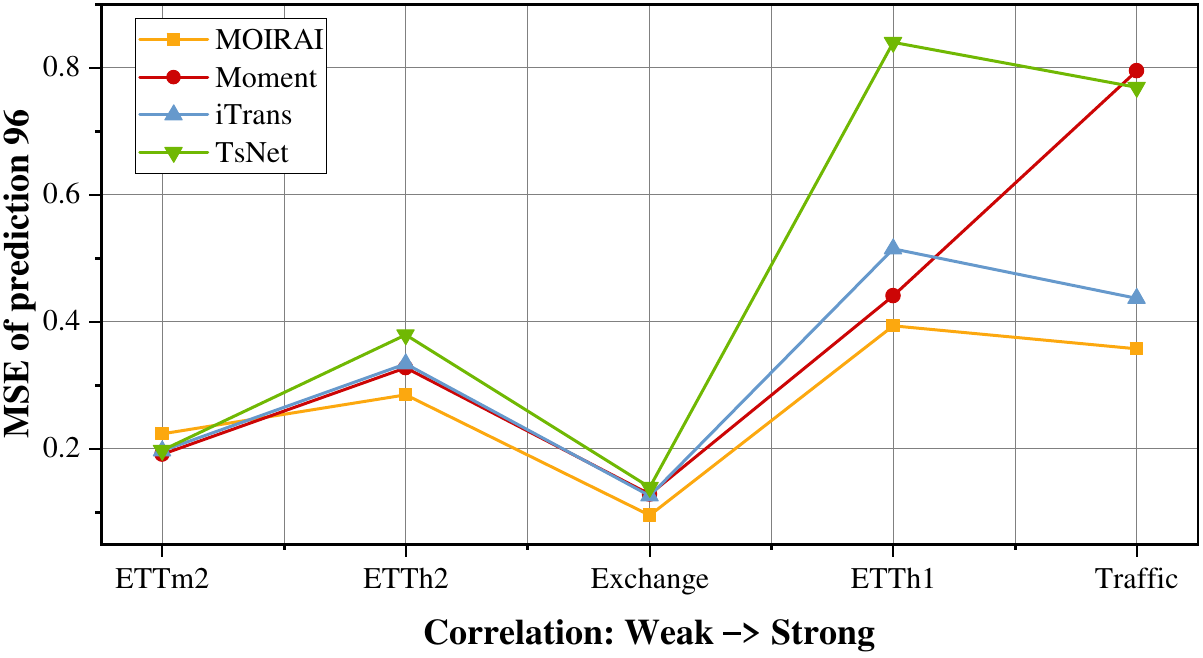}
  \caption{Method performance for varying correlation within datasets.}
\label{Correlation weak to strong}
\end{figure}

\subsection{Analysis on Different TSFMs}

\subsubsection{Channel independence vs. channel dependence}
In multivariate datasets, variables are often referred to as channels. To explore the impact of channel dependency in multivariate time series, we compare MOIRAI, Moment, iTransformer, and TimesNet across 21 datasets with varying degrees of correlations, ranging from weak to strong. We present MSE results in Figure~\ref{Correlation weak to strong}. 
The results indicate that on datasets with weak correlation, the performance of the four models is comparable. On datasets with strong correlation, it is suggested that the error rate of channel independence models (e.g., Moment) increases linearly, while the model considering channel dependence (e.g., MOIRAI) remains stable within a certain error range and outperforms other models. 
It is interesting that although both iTransformer and TimesNet consider channel dependence, their performance is far different.
This calls for models that use more appropriate way of modeling correlations.

\subsubsection{Performance on different data characteristics}
\begin{figure}[h!]
  \centering
  \includegraphics[width=0.98\linewidth]{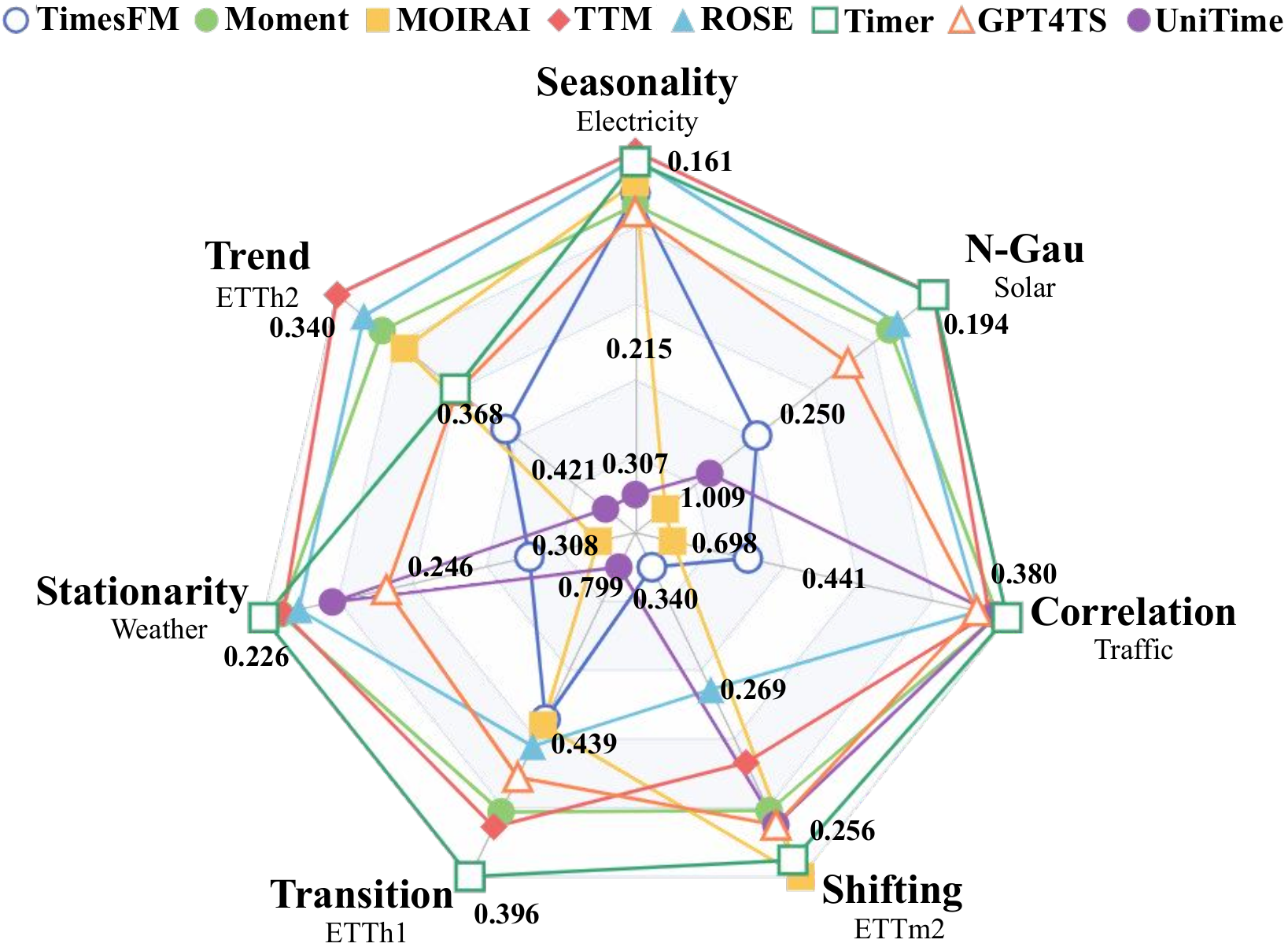}
  \caption{MSE evaluation results of TSFMs under 5\% few-shot scenario for seven key time-series characteristics.}
\label{MSE results of TSFMs across seven characteristics.}
\end{figure}

We evaluate the performance of TSFMs across different characteristics. We first score the time series datasets with respect to the above seven characteristics. For each characteristic, we select the dataset with the highest score to represent it. We present the 5\% few-shot MSE results for the models in Figure~\ref{MSE results of TSFMs across seven characteristics.}. Results reveal that no single TSFM excels across all characteristics. Notably, ROSE demonstrates exceptional performance on datasets with strong seasonality (Electricity). 
Meanwhile, Timer achieves optimal performance on datasets with strong correlation (Traffic), pronounced non-gaussianity (Solar), and most stationary (Weather). 
Similarly, TTM stands out for its performance on time series with significant trends (ETTh2). 
MOIRAI gets the best on experiences severe shifting (ETTm2).

\subsubsection{Comparison among different architectures}
\begin{figure}[t]
  \centering
  \includegraphics[width=1\linewidth]{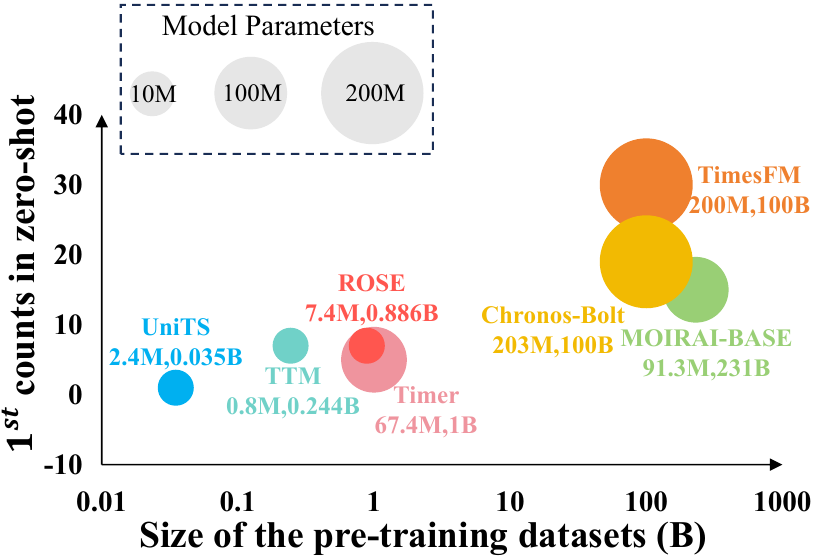}
  \caption{A comparison of the parameter counts and pre-training dataset sizes of pre-trained models, along with their zero-shot performance.}
\label{win}
\end{figure}

From Figure~\ref{win}, we observe that TimesFM, a Transformer-based model with the large parameter counts, demonstrates strong performance. 
Surprisingly, TTM, which uses a simple multi-layer perceptron (MLP) architecture, has the fewest parameters, achieves the better results than some TSFMs (e.g., UniTS and Timer). 
This phenomenon indicates that current architectures do not fully reflect the ``scaling law'', and existing TSFMs do not necessarily show a positive correlation between model parameters and performance. 
Therefore, while TimesFM performs better than most, its increase in parameters is not the only path to improving performance. These findings reveal significant opportunities for advancing research in TSFMs. In future research, we need to dive deeper into model architecture design to find a better trade-off between performance and parameter counts.

\subsubsection{Pretrain vs. No Pretrain}

To assess the practical benefits of pre-trained knowledge derived from multi-domain time series data and large-scale text data for downstream TSF tasks, we conduct comprehensive few-shot experiments (5\% training data) on established benchmark, including the ETTh2 and Weather datasets.
Our evaluation encompasses both TS pre-trained models and LLM-based models, comparing performance under two parameter initialization schemes, including leveraging pre-trained parameters and random initialization.
For LLM-based models, we randomly initialize both the LLM backbone and the additional components to ensure fair comparison without leveraging pretrained knowledge. 
Table \ref{sample-table} shows the main results: 
1) All TS pre-trained models with loaded parameters achieve significant improvements compared to random initialization.
The results indicate that pre-trained models significantly benefit from the knowledge obtained from multi-domain time series data, demonstrating their strong generalization capabilities. 
2) Conversely, most LLM-based models demonstrate degraded forecasting performance when initialized with pre-trained parameters, indicating that knowledge transfer from textual pre-training may adversely affect TSF tasks. This finding underscores the necessity for further optimization and redesign of LLM-based models are crucial to effectively leverage their potential.
3) Comparing the performance of random initialized LLM-based models with TS pre-trained models, we observe that LLM-based models achieve comparable or superior performance. This indicates that the architectural framework of LLM may possess inherent advantages for time series forecasting tasks. That means the significant potential on TSF tasks for developing effective multi-domain pre-training approaches based on LLM architectures.
\begin{table}[t]
\centering
\caption{The results of loading pre-trained parameters (denotes as "p") and random initialization (denotes as "w/o p") for TSFMs in 5\% few-shot scenario.} 
\resizebox{1\columnwidth}{!}{
\begin{tabular}{c|cc|cc|cc|cc}
\toprule
\multicolumn{1}{c|}{\multirow{3}[2]{*}{\textbf{Model}}} & \multicolumn{4}{c|}{\textbf{ETTh2}} & \multicolumn{4}{c}{\textbf{Weather}} \\  \cmidrule{2-9} 
& \multicolumn{2}{c|}{\textbf{p}} & \multicolumn{2}{c|}{\textbf{w/o p}} & \multicolumn{2}{c|}{\textbf{P}} & \multicolumn{2}{c}{\textbf{w/o p}} \\
\addlinespace\cline{2-9} \addlinespace
\multicolumn{1}{c|}{} & \textbf{MAE} & \textbf{MSE} & \textbf{MAE} & \textbf{MSE} & \textbf{MAE} & \textbf{MSE} & \textbf{MAE} & \textbf{MSE} \\
\addlinespace\cline{1-9} \addlinespace
Chronos & 0.328 & 0.292 & 0.415 & 0.371 & 0.216 & 0.183 & 0.325 & 0.273 \\
TimesFM & 0.346 & 0.292 & 0.391 & 0.332 & 0.225 & 0.182 & 0.337 & 0.311 \\
Timer & 0.346 & 0.295 & 0.422 & 0.382 & 0.194 & 0.147 & 0.331 & 0.283 \\
UniTS & 0.366 & 0.314 & 0.412 & 0.397 & 0.213 & 0.161 & 0.233 & 0.182 \\
TTM & 0.331 & 0.273 & 0.412 & 0.367 & 0.242 & 0.158 & 0.257 & 0.197 \\
Moment & 0.377 & 0.328 & 0.442 & 0.440 & 0.239 & 0.182 & 0.240 & 0.182 \\
MOIRAI & 0.329 & 0.285 & 0.417 & 0.376 & 0.220 & 0.206 & 0.329 & 0.277 \\
ROSE & 0.332 & 0.272 & 0.354 & 0.309 & 0.205 & 0.159 & 0.225 & 0.179  \\
\midrule
GPT4TS & 0.377 & 0.322 & 0.368 & 0.314 & 0.244 & 0.187 & 0.222 & 0.169 \\
S$^{2}$IP-LLM & 0.415 & 0.366 & 0.392 & 0.345 & 0.228 & 0.171 & 0.227 & 0.175 \\
Time-LLM & 0.381 & 0.342 & 0.369 & 0.314 & 0.220 & 0.170 & 0.219 & 0.165 \\
UniTime & 0.390 & 0.358 & 0.397 & 0.353 & 0.239 & 0.184 & 0.211 & 0.158 \\
LLMMixer & 0.374 & 0.315 & 0.814 & 1.286 & 0.215 & 0.162 & 0.321 & 0.275 \\
CALF & 0.362 & 0.302 & 0.403 & 0.351 & 0.217 & 0.163 & 0.216 & 0.164 \\
\bottomrule
\end{tabular}}
\label{sample-table}
\end{table}

\subsubsection{Effectiveness of lookback lengths}
\begin{figure}[t]
  \centering
  \includegraphics[width=1\linewidth]{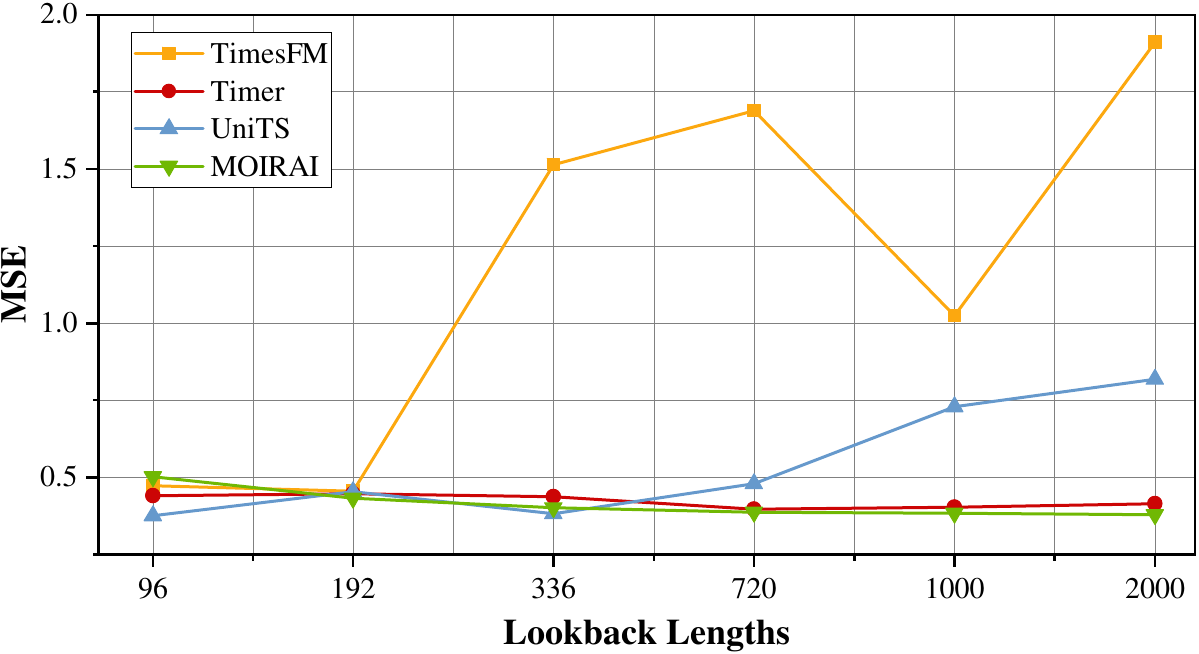}
  \caption{The MSE zero-shot results for predicting 96 length on the ETTh1 dataset with different lookback lengths.}
\label{Flexibility left}
\end{figure}

we conducted an analysis experiment to investigate whether the lookback lengths (i.e., the amount of historical information received by the model) affect model performance. Figure~\ref{Flexibility left} illustrates that MOIRAI demonstrates consistent performance gains with increasing lookback lengths, while competing models show unstable improvement patterns and occasional significant performance degradation. This suggests that when designing models, we should ensure that they can flexibly handle varying lookback lengths and effectively utilize more historical information.

\subsubsection{Model efficiency analysis}

\begin{figure}[t]
  \centering
  \includegraphics[width=1\linewidth]{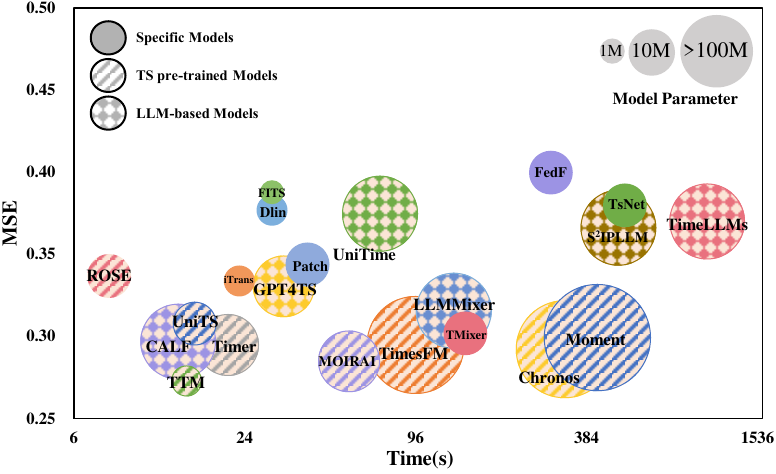}
  \caption{Comparison of parameter counts and run-time for different models.}
\label{fig: model effiency}
\end{figure}

Model efficiency is a key criterion for assessing whether TSFMs can adapt effectively to new tasks or datasets.
To assess the relationship between model efficiency and prediction accuracy of various models, we select the ETTh2 dataset and conduct comparisons between the TSFMs for 5\% few-shot and time series specific models for full-shot.
Specifically, we compare models based on three aspects: run-time (training time and inference time), number of model parameters, and prediction accuracy. As illustrated in Figure \ref{fig: model effiency}, most TS pre-trained models outperform LLM-based models in terms of running time, number of parameters, and prediction accuracy. By comparing with specific models, TSFMs exhibit varied performance levels. For instance, ROSE and TTM demonstrate superior running efficiency and prediction accuracy compared to most specific models. 
Additionally, models like S$^{2}$IP-LLM lag behind specific models in both runtime and prediction accuracy. 
This suggests that it is necessary to take model efficiency into consideration when designing a TSFM.

\subsection{Takeaways}

Based on our benchmark and analysis with \textit{TSFM-Bench}, we summarize some takeaways considering the following critical questions related to TSFMs. 

\textit{\textbf{Do TSFMs outperform specific models?}} 
Current TSFMs, especially those TS pre-trained models, exhibit superior zero-shot and few-shot learning abilities compared with specific models, which indicates their advantages in data insufficient scenarios. 
Notably, the zero-shot performance of TS pre-trained models surpassed the full-shot performance of specidic models on 10 out of the 21 datasets. When provided with more training data, TSFMs (i.e., TS pre-train models and LLM-based models) achieved optimal or suboptimal performance on 16 datasets. However, the performance of some TS pre-trained models showed no significant improvement --- even remained unchanged or exhibited notable declines --- with increases in training data. Besides, while LLM-based models underperformed compared to both TS pre-trained models and specific models, their performance demonstrated substantial growth when trained on more data. Based on these findings, we propose that future research on TSFMs should not only focus on enhancing their generalization to new data but also prioritize improving model fine-tuning capabilities and unlocking the potential of LLM.

\textit{\textbf{Which TSFMs are better?}} 
The advantages of different TSFMs depend on diverse aspects of evaluation, and no single model dominates across these aspects. 
1) Considering the two types of TSFMs, current TS pre-trained models exhibit better overall performance than LLM-based models. 
2) Different TSFMs show their advantages in dealing with datasets from diverse domains or with diverse characteristics. 
3) Large-scale TS pre-trained models such as TimesFM show the best zero-shot performance, but the situation changes given few-shot data for fine-tuning. 
4) The scaling law does not hold strictly in current TSFMs, and some small-size models such as ROSE and TTM achieve a better balance between precision and efficiency. 
It calls for benchmarks such as \textit{TSFM-Bench} to provide comprehensive evaluations of TSFMs to answer this question.

\textit{\textbf{What improvements are needed for TSFMs?}} 
1) More universal capabilities for diverse datasets and scenarios: Considering that no TSFMs wins all situations, a meaningful goal is to explore a more universal model for TSF to handle diverse forecasting situations simultaneously. From the comparison between TSFMs and specific models, the development of more powerful TSFMs should not only consider enhancing generalization performance but also increasing the upper bound forecasting performance given more training data. 
2) Better designs for utilizing large-scale pre-training knowledge: Proper training data, architecture, and pre-training strategies need to be investigated to make time series models truly take advantage of the scaling law. Especially for LLM-based models, more in-depth analysis should be made to fully extract and adapt LLM knowledge for TSF tasks.
3) Consider channel dependence: Since multivariate appears as a common characteristic in time series, how to embed generalizable correlation modeling in TSFMs from large-scale data remains an open problem.
4) More efficient training and inference: 
Given that specific models are easy to train, more efficient TSFMs that balance performance and cost are valuable for improving their practicality in real-world applications.

\section{Conclusion}

Time Series Foundation Models (TSFMs) have recently gained significant attention due to their impressive generalization capabilities in zero-shot and few-shot forecasting scenarios, leading to a surge of innovative models. This paper presents \textit{TSFM-Bench}, the first comprehensive benchmark specifically designed to quantitatively evaluate TSFMs. \textit{TSFM-Bench} encompasses a diverse array of state-of-the-art models and includes three experimental scenarios: zero-shot, few-shot, and full-shot. Using \textit{TSFM-Bench}, we thoroughly assess 14 TSFMs, revealing their strengths and weaknesses. Furthermore, we identify the inherent limitations of existing TSFMs and outline key directions for future TSFMs design. Overall, \textit{TSFM-Bench} and our evaluation provide researchers with improved tools to facilitate the development of next-generation TSFMs.

\begin{acks}
This work was partially supported by National Natural Science Foundation of China (62372179, 62406112, and 62472174). 
\end{acks}

\clearpage

\bibliographystyle{ACM-Reference-Format}
\balance
\bibliography{sample-base}

\appendix
\section{Experimental details}
\subsection{Implementation Details} 
All experiments are conducted using PyTorch~\cite{paszke2019pytorch} in Python 3.10 and execute on an NVIDIA Tesla-A800 GPU. The training process is guided by the L2 loss, employing the ADAM~\cite{kingma2014adam} optimizer. Initially, the batch size is set to 64, with the option to reduce it by half in case of an Out-Of-Memory (OOM) situation. 
The initial learning rate is set to 0.0001 and dynamically adjusted through a simulated annealing approach over a total of 20 training epochs. Additionally, we employed early stopping (patience=3) to avoid overfitting.

\section{Datasets}
\label{appendix DATASETS}

\subsection{Datasets Collection}
\label{Datasets Collection}
We collects a set of publicly available multivariate time series datasets that cover a wide range of domains and characteristics, including various frequencies (e.g., min, hour, day, month), different lengths ranging from 728 to 57,600. 
\begin{itemize}[left=0.1cm]
\item Electricity: ETT~\cite{zhou2021informer} contain 7 variates collected from two different electric transformers from July 2016 to July 2018. 
Electricity~\cite{misc_electricityloaddiagrams20112014_321} contains the electricity consumption of 321 customers from 2016 to 2019. 
\item Traffic: Traffic~\cite{wu2021autoformer} contains road occupancy rates measured by 862 sensors on freeways in the San Francisco Bay Area from 2015 to 2016. 
PEMS08~\cite{song2020spatial} records traffic flow time series collected from the CalTrans PeMS.
\item Energy: Solar~\cite{lai2018modeling} records solar power generation from 137 PV plants in 2006, every 10 minutes. 
Wind~\cite{li2022generative} contains energy production data from wind farms in Australia. Original data was collected at 15 minute frequencey.
\item Environment: Weather~\cite{wu2021autoformer} collects 21 meteorological indicators, including temperature and barometric pressure, for Germany in 2020, recorded every 10 minutes. 
AQShunyi~\cite{zhang2017cautionary} is a air quality datasets from a measurement station, over a period of 4 years. 
\item Economic: Exchange~\cite{lai2018modeling} collects the daily exchange rates of eight countries. 
FRED-MD~\cite{mccracken2016fred} contains monthly macro-economic indicators from the Federal Reserve Bank. 
\item Nature: ZafNoo~\cite{poyatos2020global} is collected from the Sapflux data project and includes sap flow measurements and environmental variables. 
CzeLan~\cite{poyatos2020global} from the Sapflux data project includes sap flow measurements and nvironmental variables.
\item Heathy: ILI~\cite{wu2021autoformer} records indicators of patients data from Centers for Disease Control and Prevention. 
Covid-19~\cite{panagopoulos2021transfer} records the prevalence in different regions.
\item Stock: NASDAQ~\cite{feng2019temporal} and NYSE~\cite{feng2019temporal} record the historical value of stock changes.
\item Banking: NN5~\cite{taieb2012review} is the daily records from ATMs in the UK. 
\item Web: Wike2000~\cite{gasthaus2019probabilistic} is the daily page views of Wikipedia pages.
\end{itemize}

\subsection{Time series characteristics}
\label{characteristic formula}

\subsubsection{\textbf{Non-Gaussianity (N-Gau)}}
Non-Gaussianity complexity refers to the extent to which the distribution of values within a time series segment deviates from a Gaussian distribution, measuring the intricacy and variability of the data distribution. Algorithm~\ref{non-Gaussianity} details the calculation process.

\begin{algorithm}[h]
\caption{Calculating Non-Gaussianity of Time Series}
\footnotesize
\begin{flushleft}
{\bf Input:} 
Time series $X\in \mathbb{R}^{T\times 1}$, window length $w$

{\bf Output:}
Average non-Gaussianity $\mathit{avg\_JSD}$ of $X$
\end{flushleft}
\begin{algorithmic}[1]
\State {\bf function} $\mathit{JSD}(P, Q)$
\State \hspace{0.1in} $M \leftarrow 0.5 \times (P + Q)$
\State \hspace{0.1in} $\mathit{kl\_p\_m} \leftarrow \mathit{KL\_divergence}(P, M)$
\State \hspace{0.1in} $\mathit{kl\_q\_m} \leftarrow \mathit{KL\_divergence}(Q, M)$
\State \hspace{0.1in} {\bf return} $0.5 \times (\mathit{kl\_p\_m} + \mathit{kl\_q\_m})$
\State {\bf end function}
\State Divide $X$ into windows $P_1, P_2, \ldots, P_n$ where $P_i \in \mathbb{R}^{w \times 1}$
\State Initialize $\text{total\_JSD} \leftarrow 0$
\State {\bf for} each window $P_i$ {\bf do}
\State \hspace{0.1in} Fit a Gaussian distribution $Q_i$ to $P_i$
\State \hspace{0.1in} Calculate the JS Divergence $\text{JSD}(P_i, Q_i)$
\State \hspace{0.1in} $\text{total\_JSD} \leftarrow \text{total\_JSD} + \text{JSD}(P_i, Q_i)$
\State {\bf end for}
\State $\text{avg\_JSD} \leftarrow \frac{\text{total\_JSD}}{n}$
\State {\bf return} $\text{avg\_JSD}$
\end{algorithmic}
\label{non-Gaussianity}
\end{algorithm}

\subsubsection{\textbf{Stationarity}}
Stationarity refers to the mean of any observation in a time series $X=\langle x_1,x_2,...,x_n\rangle$ is constant, and the variance is finite for all observations. Algorithm~\ref{Stationarity} details the calculation process.

\begin{algorithm}[!]
\caption{Calculating Stationarity Values of Time Series}
\footnotesize
\begin{flushleft}
{\bf Input:} 
Time series $X\in \mathbb{R}^{T\times 1}$

{\bf Output:}
Stationarity value $\gamma$ ~$\in\{0,1\}$ of $X$
\end{flushleft}
\begin{algorithmic}[1]
\State $ s \leftarrow \mathit{ADF}(X)$
\State {\bf return} $\gamma \leftarrow (s <= 0.05)$
\end{algorithmic}
\label{Stationarity}
\end{algorithm}

\subsubsection{\textbf{Trend}}
The trend of a time series reflects its long-term movement or pattern, indicating the overall direction of the data over time. Trend Strength can be defined as in Algorithm~\ref{Trend}. 

\begin{algorithm}[h]
\caption{Calculating Trend Values of Time Series}
\footnotesize
\begin{flushleft}
{\bf Input:} 
Time series $X\in \mathbb{R}^{T\times 1}$

{\bf Output:}
$\mathit{Trend\_Strength}$ $\beta \in(0,1)$ of $X$
\end{flushleft}
\begin{algorithmic}[1]
\State $ S, T, R \leftarrow \mathit{STL}(X); X = S + T + R $

\State {\bf return} $\beta \leftarrow \mathit{max} \left(0, 1 - \frac{\mathit{var}\left(R\right)}{\mathit{var}\left(T + R\right)}\right)$
\end{algorithmic}
\label{Trend}
\end{algorithm}

\subsubsection{\textbf{Seasonality}}
Seasonality refers to the phenomenon where changes in a time series repeat at specific intervals. Algorithm~\ref{Seasonality} details the calculation process.

\begin{algorithm}[h]
\caption{Calculating Seasonality Values of Time Series}
\footnotesize
\begin{flushleft}
{\bf Input:} 
Time series $X\in \mathbb{R}^{T\times 1}$

{\bf Output:}
$ \mathit{Seasonality\_Strength} $ $\zeta \in(0,1)$ of $X$
\end{flushleft}
\begin{algorithmic}[1]
\State $ S, T, R \leftarrow \mathit{STL}(X); X = S + T + R $
\State {\bf return} $\zeta \leftarrow \max \left(0, 1-\frac{\mathit{var}\left(R\right)}{\mathit{var}\left(S+R\right)}\right)$
\end{algorithmic}
\label{Seasonality}
\end{algorithm}

\subsubsection{\textbf{Shifting}}
Shifting occurs when the probability distribution of a time series changes over time, caused by system dynamics or random events. Algorithm~\ref{Shifting} details the calculation process.

\begin{algorithm}[!h]
\caption{Calculating Shifting Values of Time Series}
\footnotesize
\begin{flushleft}
{\bf Input:} 
Time series $X\in \mathbb{R}^{T\times 1}$

{\bf Output:}
Shifting value $\delta$ ~$\in(0,1)$ of $X$
\end{flushleft}
\begin{algorithmic}[1]
\State Normalize $X$ by calculating the z-score to obtain $Z\in \mathbb{R}^{T\times 1}$
\State $Z_{\min} \leftarrow \min(Z),~Z_{\max} \leftarrow \max(Z)$
\State $S \leftarrow \{s_{i}~|~s_{i} \leftarrow Z_{\min}+(i-1)\frac{Z_{\max}-Z_{\min}}{m},1\leq i\leq m\}$ where m is the number of thresholds
\State {\bf for} $s_i$ in $S$ {\bf do}
\State \hspace{0.1in} $K \leftarrow \{j~|~ Z_j>s_{i},1\leq j\leq T\}$,~$M_i$ $\leftarrow$ $\mathit{median}(K)$,~ $1 \leq i \leq m$
\State {\bf end for}
\State  $M ^\prime \leftarrow$ $\mathit{Min}$--$\mathit{Max}$ $\mathit{Normalization}(M)$
\State {\bf return} $\text{$\delta$} \leftarrow \mathit{abs(\mathit{median}(\{M_1 ^\prime, M_2^\prime, ..., M_m^\prime\}))}$
\end{algorithmic}
\label{Shifting}
\end{algorithm}

\subsubsection{\textbf{Transition}}
Transition refers to the trace of the covariance of transition
matrix between symbols in a 3-letter alphabet~\cite{lubba2019catch22}. Algorithm~\ref{Transition} details the calculation process. 
\begin{algorithm}[!h]
\caption{Calculating Transition Values of Time Series}
\footnotesize
\begin{flushleft}
{\bf Input:} 
Time series $X\in \mathbb{R}^{T\times 1}$

{\bf Output:}
Transition value $\Delta$ ~$\in(0, \frac{1}{3})$ of $X$
\end{flushleft}
\begin{algorithmic}[1]
\State Calculate the first zero crossing of the autocorrelation function: 
\Statex $\tau \leftarrow \mathit{firstzero\_ac}(X)$  
\State Generate $Y\in \mathbb{R}^{T^\prime\times 1}$ by downsampling X with stride $\tau$
\State Define index  $I = \mathit{argsort}(Y) \in \mathbb{R}^{T^\prime\times 1}$, then characterize  $Y$ to obtain $Z\in \mathbb{R}^{T^\prime\times 1}$:
\State {\bf for} $j \in [0: T^\prime]$ {\bf do}
\State \hspace{0.1in} $Z[j] \leftarrow \mathit{floor}(~I[j] / \frac{1}{3}T^\prime)$
\State {\bf end for}
\State Generate a transition matrix $M\in \mathbb{R}^{3\times 3}$:
\State {\bf for} $j \in [0: T^\prime] $ {\bf do}
\State \hspace{0.1in} $M[Z[j]-1][Z[j+1]-1]$++
\State {\bf end for}
\State $M^\prime \leftarrow \frac{1}{T^\prime}M$
\State Compute the covariance matrix $C$ between the columns of $M^\prime$
\State {\bf return} $\Delta \leftarrow \mathit{tr}(C)$
\end{algorithmic}
\label{Transition}
\end{algorithm}

\subsubsection{\textbf{Correlation}}
Correlation refers to the possibility that different variables in a multivariate time series may share common trends or patterns. Algorithm~\ref{Correlation} details the calculation process.

\begin{algorithm}[!h]
\caption{Calculating Correlation Values of Time Series}
\footnotesize
\begin{flushleft}
{\bf Input:} 
Time series $X\in \mathbb{R}^{T\times N}$

{\bf Output:}
Correlation value $\Delta$ ~$\in(0, 1)$ of $X$
\end{flushleft}
\begin{algorithmic}[1]
\State Get the representation for each channel using the Catch22 library: 
\Statex $F=\langle F^1,F^2,...,F^N\rangle\in \mathbb{R}^{22\times N}$ $\leftarrow \mathit{Catch22}(X)$  
\State Calculate the Pearson correlation coefficients between all pairs of channels:
\Statex $P=\left\{r(F^{i},F^{j})\mid 1\leq i\leq N,i+1\leq j\leq N,i,j\in N^{\ast}\right\}$
\State Compute the correlation by computing the mean and variance of all Pearson correlation coefficients (PCCs)
\Statex $\mathit{Correlation}=\mathit{mean}\left(P\right)+\frac{1}{1+\mathit{var}\left(P\right)}$
\State {\bf return} $Correlation$
\end{algorithmic}
\label{Correlation}
\end{algorithm}

\section{Full results}
\label{Full results}

The full results are in Table~\ref{tab:zero-shot(mse)}, Table~\ref{tab:zero-shot(mae)}, Table~\ref{tab:few-shot(mse)}, Table~\ref{tab:few-shot(mae)}, Table~\ref{tab:full-shot(mse)} and Table~\ref{tab:full-shot(mae)}.

\begin{table*}[!h]
\small
\centering
\caption{Pre-trained model results in the \textbf{zero-shot setting}. The results are \textbf{MSE} of each prediction length.}
\resizebox{2\columnwidth}{!}{
\begin{tabular}{c|c|ccccccccccccccccccccc}
\toprule
\textbf{Model} & \textbf{Pred} & \textbf{ETTh1} & \textbf{ETTh2} & \textbf{ETTm1} & \textbf{ETTm2} & \textbf{Electricity} & \textbf{Traffic} & \textbf{PEMS08} & \textbf{Solar} & \textbf{Wind} & \textbf{Weather} & \textbf{AQShunyi} & \textbf{Exchange} & \textbf{FRED-MD} & \textbf{ZafNoo} & \textbf{CzeLan} & \textbf{ILI} & \textbf{Covid-19} & \textbf{NASDAQ} & \textbf{NYSE} & \textbf{NN5} & \textbf{Wike2000} \\
\midrule
\multirow{4}{*}{Chronos} & 96 & 0.388 & 0.292 & \textbf{0.339} & \textbf{0.181} & \uline{0.131} & 0.401 & 0.456 & 0.430 & 1.130 & 0.183 & 0.702 & 0.138 & \textbf{27.269} & \uline{0.455} & 0.266 & 2.498 & \textbf{0.991} & \textbf{0.485} & \uline{0.251} & 1.096 & \uline{445.311} \\
 & 192 & 0.440 & 0.362 & 0.392 & 0.253 & 0.149 & 0.427 & 0.648 & \uline{0.396} & 1.300 & 0.227 & 0.771 & 0.261 & 47.583 & 0.524 & 0.288 & 2.810 & \uline{1.276} & 0.746 & 0.461 & 1.097 & 514.033 \\
 & 336 & 0.477 & 0.404 & 0.440 & 0.318 & 0.167 & 0.448 & 0.747 & 0.409 & 1.451 & 0.286 & 0.817 & 0.451 & 75.692 & 0.568 & 0.319 & 2.950 & 1.596 & 1.071 & 0.689 & 1.109 & 528.929 \\
 & 720 & 0.475 & 0.412 & 0.530 & 0.417 & 0.216 & 0.516 & 0.938 & 0.453 & 1.604 & 0.368 & 0.863 & 1.105 & 106.614 & 0.678 & 0.385 & 3.201 & 1.963 & 1.231 & 0.954 & 1.151 & 559.220 \\
\midrule
\multirow{4}{*}{TimesFM} & 96 & 0.387 & 0.327 & \uline{0.357} & 0.205 & \textbf{0.119} & \textbf{0.327} & 1.131 & \textbf{0.352} & 1.229 & \textbf{0.122} & 0.807 & 0.110 & \uline{27.722} & 0.518 & \textbf{0.213} & \textbf{1.725} & 10.362 & 0.664 & \textbf{0.235} & 0.830 & \textbf{420.043} \\
 & 192 & 0.444 & 0.400 & 0.412 & 0.294 & 0.137 & \uline{0.354} & 1.609 & 0.399 & 1.503 & \uline{0.169} & 0.868 & 0.220 & 44.831 & 0.606 & \uline{0.263} & \uline{1.773} & 14.376 & 1.032 & 0.472 & 0.774 & 472.763 \\
\bottomrule
\end{tabular}

}

\label{tab:zero-shot(mse)}
\end{table*}

\begin{table*}[!h]
\small
\centering
\caption{Pre-trained model results in the \textbf{zero-shot setting}. The results are \textbf{MAE} of each prediction length.}
\resizebox{2\columnwidth}{!}{
\begin{tabular}{c|c|ccccccccccccccccccccc}
\toprule
\textbf{Model} & \textbf{Pred} & \textbf{ETTh1} & \textbf{ETTh2} & \textbf{ETTm1} & \textbf{ETTm2} & \textbf{Electricity} & \textbf{Traffic} & \textbf{PEMS08} & \textbf{Solar} & \textbf{Wind} & \textbf{Weather} & \textbf{AQShunyi} & \textbf{Exchange} & \textbf{FRED-MD} & \textbf{ZafNoo} & \textbf{CzeLan} & \textbf{ILI} & \textbf{Covid-19} & \textbf{NASDAQ} & \textbf{NYSE} & \textbf{NN5} & \textbf{Wike2000} \\
\midrule
\multirow{4}{*}{Chronos} & 96 & \textbf{0.387} & \textbf{0.328} & \textbf{0.340} & \textbf{0.248} & \uline{0.218} & 0.249 & 0.399 & 0.336 & 0.690 & 0.216 & 0.482 & 0.258 & \uline{0.877} & \textbf{0.382} & \uline{0.270} & 0.940 & \textbf{0.041} & \textbf{0.481} & \textbf{0.308} & 0.732 & 1.092 \\
 & 192 & 0.416 & 0.371 & 0.372 & 0.296 & 0.235 & 0.260 & 0.463 & \uline{0.325} & 0.783 & 0.258 & 0.506 & 0.362 & 1.175 & 0.424 & 0.284 & 1.024 & \uline{0.047} & 0.612 & 0.437 & 0.756 & 1.228 \\
 & 336 & 0.434 & 0.403 & 0.398 & 0.337 & 0.254 & 0.269 & 0.483 & 0.336 & 0.856 & 0.297 & 0.523 & 0.485 & 1.486 & 0.450 & 0.306 & 1.032 & 0.054 & 0.736 & 0.548 & 0.773 & 1.275 \\
 & 720 & 0.447 & 0.420 & 0.442 & 0.396 & 0.294 & 0.298 & 0.561 & 0.362 & 0.922 & 0.348 & 0.548 & 0.788 & 1.767 & 0.498 & 0.350 & 1.029 & 0.062 & 0.791 & 0.669 & 0.799 & 1.338 \\
\midrule
\multirow{4}{*}{TimesFM} & 96 & 0.398 & 0.352 & \uline{0.367} & \uline{0.267} & \textbf{0.212} & 0.220 & 0.759 & \textbf{0.302} & 0.722 & \textbf{0.158} & 0.513 & 0.233 & \textbf{0.836} & 0.413 & \textbf{0.229} & \textbf{0.806} & 0.060 & 0.546 & \uline{0.322} & 0.609 & \textbf{0.935} \\
 & 192 & 0.429 & 0.398 & 0.402 & 0.320 & 0.229 & 0.235 & 0.944 & 0.329 & 0.835 & \uline{0.205} & 0.544 & 0.334 & 1.079 & 0.465 & 0.271 & \uline{0.853} & 0.076 & 0.695 & 0.467 & \textbf{0.600} & \uline{1.040} \\
 & 336 & 0.451 & 0.432 & 0.426 & 0.365 & 0.249 & 0.248 & 0.939 & 0.362 & 0.925 & 0.267 & 0.564 & 0.456 & 1.332 & 0.498 & 0.320 & 0.875 & 0.093 & 0.754 & 0.591 & \uline{0.602} & 1.108 \\
\bottomrule
\end{tabular}

}

\label{tab:zero-shot(mae)}
\end{table*}

\begin{table*}[b]
\small
\centering
\caption{Model results in the \textbf{5\% few-shot setting}. The results are \textbf{MSE} of each prediction length.}
\resizebox{2\columnwidth}{!}{
\begin{tabular}{c|c|cccccccc|cccccc|ccccccc}
\toprule
\textbf{Dataset} & \textbf{Pred} & \textbf{Chronos} & \textbf{TimesFM} & \textbf{Moment} & \textbf{UniTS} & \textbf{MOIRAI} & \textbf{TTM} & \textbf{ROSE} & \textbf{Timer} & \textbf{LLMMixer} & \textbf{S$^2$IPLLM} & \textbf{CALF} & \textbf{Time-LLM} & \textbf{GPT4TS} & \textbf{UniTime} & \textbf{Patch} & \textbf{Dlin} & \textbf{FITS} & \textbf{iTrans} & \textbf{FedF} & \textbf{TsNet} & \textbf{TMixer} \\
\midrule
\multirow{4}{*}{ETTh1} & 96 & 0.388 & 0.387 & 0.408 & 0.382 & 0.394 & \uline{0.363} & 0.378 & \textbf{0.359} & 0.460 & 0.665 & 0.413 & 0.539 & 0.438 & 0.716 & 0.404 & 0.391 & 0.755 & 0.515 & 0.476 & 0.840 & 0.449 \\
 & 192 & 0.440 & 0.436 & 0.428 & 0.431 & 0.430 & \textbf{0.388} & 0.415 & \uline{0.399} & 0.481 & 0.676 & 0.426 & 0.702 & 0.460 & 0.801 & 0.431 & 0.431 & 0.669 & 0.534 & 0.533 & 0.817 & 0.743 \\
 & 336 & 0.477 & 0.477 & 0.464 & 0.476 & 0.450 & \textbf{0.405} & \uline{0.414} & 0.422 & 0.555 & 0.672 & 0.443 & 0.716 & 0.462 & 0.820 & 0.456 & 0.452 & 0.783 & 0.669 & 0.606 & 0.879 & 0.776 \\
 & 720 & 0.475 & 0.482 & 0.536 & 0.493 & 0.457 & \uline{0.428} & \textbf{0.412} & 0.453 & 0.598 & 0.717 & 0.495 & 0.739 & 0.509 & 0.860 & 0.499 & 0.628 & 0.916 & 0.571 & 0.613 & 0.882 & 0.585 \\
\midrule
\multirow{4}{*}{ETTh2} & 96 & 0.292 & 0.295 & 0.299 & 0.308 & \uline{0.285} & \textbf{0.273} & 0.298 & 0.295 & 0.315 & 0.366 & 0.297 & 0.342 & 0.331 & 0.358 & 0.344 & 0.377 & 0.388 & 0.334 & 0.400 & 0.380 & 0.302 \\
 & 192 & 0.362 & 0.380 & 0.347 & 0.385 & 0.352 & \textbf{0.336} & \uline{0.342} & 0.361 & 0.382 & 0.381 & 0.360 & 0.402 & 0.368 & 0.419 & 0.378 & 0.448 & 0.418 & 0.393 & 0.494 & 0.461 & 0.575 \\
 & 336 & 0.400 & 0.430 & 0.365 & 0.424 & 0.384 & \uline{0.364} & \textbf{0.356} & 0.396 & 0.396 & 0.392 & 0.387 & 0.388 & 0.378 & 0.430 & 0.384 & 0.518 & 0.440 & 0.445 & 0.407 & 0.448 & 0.377 \\
\bottomrule
\end{tabular}

}

\label{tab:few-shot(mse)}
\end{table*}

\begin{table*}[h]
\small
\centering
\caption{Model results in the \textbf{5\% few-shot setting}. The results are \textbf{MAE} of each prediction length.}
\resizebox{2\columnwidth}{!}{
\begin{tabular}{c|c|cccccccc|cccccc|ccccccc}
\toprule
\textbf{Dataset} & \textbf{Pred} & \textbf{Chronos} & \textbf{TimesFM} & \textbf{Moment} & \textbf{UniTS} & \textbf{MOIRAI} & \textbf{TTM} & \textbf{ROSE} & \textbf{Timer} & \textbf{LLMMixer} & \textbf{S$^2$LLM} & \textbf{CALF} & \textbf{TimeLLMs} & \textbf{GPT4TS} & \textbf{UniTime} & \textbf{Patch} & \textbf{Dlin} & \textbf{FITS} & \textbf{iTrans} & \textbf{FedF} & \textbf{TsNet} & \textbf{TMixer} \\
\midrule
\multirow{4}{*}{ETTh1} & 96 & \textbf{0.387} & 0.398 & 0.422 & 0.394 & 0.399 & \uline{0.392} & 0.398 & 0.393 & 0.463 & 0.553 & 0.417 & 0.494 & 0.445 & 0.575 & 0.410 & 0.410 & 0.590 & 0.481 & 0.475 & 0.615 & 0.458 \\
 & 192 & 0.416 & 0.427 & 0.436 & 0.422 & 0.422 & \textbf{0.407} & 0.418 & \uline{0.414} & 0.474 & 0.562 & 0.440 & 0.565 & 0.458 & 0.614 & 0.441 & 0.440 & 0.563 & 0.504 & 0.505 & 0.608 & 0.599 \\
 & 336 & 0.434 & 0.452 & 0.465 & 0.444 & 0.437 & \textbf{0.424} & \textbf{0.424} & 0.430 & 0.522 & 0.582 & 0.454 & 0.595 & 0.467 & 0.627 & 0.461 & 0.449 & 0.614 & 0.569 & 0.535 & 0.644 & 0.636 \\
 & 720 & \uline{0.447} & 0.470 & 0.514 & 0.470 & 0.458 & 0.453 & \textbf{0.442} & 0.464 & 0.554 & 0.616 & 0.494 & 0.612 & 0.511 & 0.654 & 0.498 & 0.578 & 0.678 & 0.529 & 0.559 & 0.673 & 0.546 \\
\midrule
\multirow{4}{*}{ETTh2} & 96 & \textbf{0.328} & 0.343 & 0.356 & 0.347 & \uline{0.329} & 0.331 & 0.355 & 0.346 & 0.374 & 0.415 & 0.357 & 0.381 & 0.383 & 0.390 & 0.394 & 0.426 & 0.426 & 0.377 & 0.443 & 0.405 & 0.366 \\
 & 192 & \textbf{0.371} & 0.395 & 0.387 & 0.397 & 0.374 & \textbf{0.371} & 0.388 & 0.394 & 0.412 & 0.424 & 0.395 & 0.442 & 0.406 & 0.446 & 0.415 & 0.469 & 0.446 & 0.426 & 0.499 & 0.452 & 0.534 \\
 & 336 & 0.404 & 0.437 & 0.406 & 0.430 & 0.403 & \textbf{0.397} & \uline{0.400} & 0.426 & 0.434 & 0.435 & 0.418 & 0.431 & 0.421 & 0.455 & 0.425 & 0.513 & 0.462 & 0.462 & 0.452 & 0.466 & 0.424 \\
\bottomrule
\end{tabular}

}

\label{tab:few-shot(mae)}
\end{table*}

\begin{table*}[h]
\small
\centering
\caption{Model results in the \textbf{full-shot setting}. The results are \textbf{MSE} of each prediction length.}
\resizebox{1.85\columnwidth}{!}{

\begin{tabular}{c|c|cccccc|cccc|ccccccc}
\toprule
\textbf{Dataset} & \textbf{Pred} & \textbf{Chronos} & \textbf{UniTS} & \textbf{MOIRAI} & \textbf{TTM} & \textbf{ROSE} & \textbf{Timer} & \textbf{LLMMixer} & \textbf{CALF} & \textbf{GPT4TS} & \textbf{UniTime} & \textbf{Patch} & \textbf{Dlin} & \textbf{FITS} & \textbf{iTrans} & \textbf{FedF} & \textbf{TsNet} & \textbf{TMixer} \\
\midrule
\multirow{4}{*}{ETTh1} & 96 & 0.388 & 0.434 & 0.394 & \uline{0.361} & \textbf{0.360} & 0.413 & 0.374 & 0.398 & 0.391 & 0.392 & 0.377 & 0.379 & 0.376 & 0.386 & 0.379 & 0.389 & 0.372 \\
 & 192 & 0.440 & 0.461 & 0.430 & \uline{0.393} & \textbf{0.390} & 0.487 & 0.428 & 0.440 & 0.421 & 0.487 & 0.409 & 0.408 & 0.400 & 0.424 & 0.420 & 0.440 & 0.413 \\
 & 336 & 0.477 & 0.441 & 0.450 & \uline{0.411} & \textbf{0.404} & 0.501 & 0.439 & 0.477 & 0.428 & 0.504 & 0.431 & 0.440 & 0.419 & 0.449 & 0.458 & 0.523 & 0.438 \\
 & 720 & 0.474 & 0.565 & 0.457 & \uline{0.426} & \textbf{0.418} & 0.538 & 0.488 & 0.494 & 0.459 & 0.617 & 0.457 & 0.471 & 0.435 & 0.495 & 0.474 & 0.521 & 0.486 \\
\midrule
\multirow{4}{*}{ETTh2} & 96 & 0.292 & 0.311 & 0.285 & \uline{0.270} & \textbf{0.267} & 0.324 & 0.295 & 0.319 & 0.299 & 0.299 & 0.274 & 0.300 & 0.277 & 0.297 & 0.337 & 0.334 & 0.281 \\
 & 192 & 0.362 & 0.470 & 0.352 & 0.362 & \uline{0.333} & 0.409 & 0.357 & 0.390 & 0.346 & 0.374 & 0.348 & 0.387 & \textbf{0.331} & 0.372 & 0.415 & 0.404 & 0.349 \\
 & 336 & 0.400 & 0.429 & 0.384 & 0.367 & \uline{0.355} & 0.419 & 0.382 & 0.426 & 0.368 & 0.403 & 0.377 & 0.490 & \textbf{0.350} & 0.388 & 0.389 & 0.389 & 0.366 \\
\bottomrule
\end{tabular}

}
\label{tab:full-shot(mse)}
\end{table*}

\begin{table*}[h]
\small
\centering
\caption{Model results in the \textbf{full-shot setting}. The results are \textbf{MAE} of each prediction length.}
\resizebox{1.85\columnwidth}{!}{

\begin{tabular}{c|c|cccccc|cccc|ccccccc}
\toprule
\textbf{Dataset} & \textbf{Pred} & \textbf{Chronos} & \textbf{UniTS} & \textbf{MOIRAI} & \textbf{TTM} & \textbf{ROSE} & \textbf{Timer} & \textbf{LLMMixer} & \textbf{CALF} & \textbf{GPT4TS} & \textbf{UniTime} & \textbf{Patch} & \textbf{Dlin} & \textbf{FITS} & \textbf{iTrans} & \textbf{FedF} & \textbf{TsNet} & \textbf{TMixer} \\
\midrule
\multirow{4}{*}{ETTh1} & 96 & \uline{0.387} & 0.429 & 0.399 & 0.392 & \textbf{0.386} & 0.424 & 0.404 & 0.411 & 0.411 & 0.413 & 0.397 & 0.403 & 0.396 & 0.405 & 0.419 & 0.412 & 0.401 \\
 & 192 & 0.416 & 0.448 & 0.422 & \uline{0.415} & \textbf{0.406} & 0.459 & 0.432 & 0.434 & 0.428 & 0.478 & 0.425 & 0.419 & 0.418 & 0.440 & 0.444 & 0.443 & 0.430 \\
 & 336 & 0.434 & 0.451 & 0.437 & \uline{0.429} & \textbf{0.420} & 0.471 & 0.452 & 0.454 & 0.437 & 0.489 & 0.444 & 0.440 & 0.435 & 0.460 & 0.466 & 0.487 & 0.450 \\
 & 720 & \textbf{0.446} & 0.525 & 0.458 & \uline{0.450} & \uline{0.450} & 0.505 & 0.491 & 0.482 & 0.483 & 0.552 & 0.477 & 0.493 & 0.458 & 0.487 & 0.488 & 0.495 & 0.484 \\
\midrule
\multirow{4}{*}{ETTh2} & 96 & \uline{0.328} & 0.365 & 0.329 & 0.330 & \textbf{0.327} & 0.366 & 0.354 & 0.362 & 0.359 & 0.353 & 0.337 & 0.364 & 0.345 & 0.348 & 0.380 & 0.370 & 0.351 \\
 & 192 & \textbf{0.371} & 0.453 & 0.374 & 0.384 & \textbf{0.371} & 0.410 & 0.393 & 0.412 & 0.391 & 0.399 & 0.384 & 0.423 & 0.379 & 0.403 & 0.428 & 0.413 & 0.387 \\
 & 336 & 0.404 & 0.437 & 0.403 & 0.400 & \textbf{0.392} & 0.428 & 0.413 & 0.441 & 0.410 & 0.424 & 0.416 & 0.487 & \uline{0.396} & 0.417 & 0.457 & 0.435 & 0.413 \\
\bottomrule
\end{tabular}

}
\label{tab:full-shot(mae)}
\end{table*}

\end{document}